\documentclass[acmsmall,screen]{acmart}

\AtBeginDocument{%
  \providecommand\BibTeX{{%
    \normalfont B\kern-0.5em{\scshape i\kern-0.25em b}\kern-0.8em\TeX}}}

\setcopyright{acmcopyright}
\copyrightyear{2021}
\acmYear{2021}
\acmDOI{10.1145/1122445.1122456}

\usepackage{graphicx} 

\usepackage{amssymb}
\usepackage{longtable}
\usepackage{setspace}
\usepackage{url}
\newcommand{\myparagraph}[1]{\smallskip \noindent{\bf {#1}.}}
\usepackage{bm}
\usepackage{color}
\usepackage{multirow}
\usepackage{subfigure}

\acmJournal{TIST}
\acmVolume{37}
\acmNumber{4}
\acmArticle{111}
\acmMonth{4}

\begin{document}

\title{A Survey on Text Classification: From Traditional to Deep Learning}

\author{Qian Li}
\affiliation{%
  \institution{Beihang University}
  \city{Haidian}
  \state{Beijing}
  \country{China}}
\email{liqian@act.buaa.edu.cn}

\author{Hao Peng}
\affiliation{%
  \institution{Beihang University}
  \city{Haidian}
  \state{Beijing}
  \country{China}}
\email{penghao@act.buaa.edu.cn}

\author{Jianxin Li}
\authornote{Corresponding author}
\affiliation{%
  \institution{Beihang University}
  \city{Haidian}
  \state{Beijing}
  \country{China}}
\email{lijx@act.buaa.edu.cn}

\author{Congying Xia}
\affiliation{%
  \institution{University of Illinois at Chicago}
  \city{Chicago}
  \state{IL}
  \country{USA}}
\email{cxia8@uic.edu}

\author{Renyu Yang}
\affiliation{%
  \institution{University of Leeds}
  \city{Leeds}
  \state{England}
  \country{UK}
  }
\email{r.yang1@leeds.ac.uk}

\author{Lichao Sun}
\affiliation{%
    \institution{Lehigh University}
    \city{Bethlehem}
    \state{PA}
    \country{USA}
}
\email{james.lichao.sun@gmail.com}

\author{Philip S. Yu}
\affiliation{%
  \institution{University of Illinois at Chicago}
  \city{Chicago}
  \state{IL}
  \country{USA}
  }
\email{psyu@uic.edu}

\author{Lifang He}
\affiliation{%
    \institution{Lehigh University}
    \city{Bethlehem}
    \state{PA}
    \country{USA}
}
\email{lih319@lehigh.edu}

\renewcommand{\shortauthors}{Qian Li, et al.}

\begin{abstract}

Text classification is the most fundamental and essential task in natural language processing. 
The last decade has seen a surge of research in this area due to the unprecedented success of deep learning. 
Numerous methods, datasets, and evaluation metrics have been proposed in the literature, raising the need for a comprehensive and updated survey. 
This paper fills the gap by reviewing the state-of-the-art approaches from 1961 to 2021, focusing on models from traditional models to deep learning. 
We create a taxonomy for text classification according to the text involved and the models used for feature extraction and classification.
We then discuss each of these categories in detail, dealing with both the technical developments and benchmark datasets that support tests of predictions. 
A comprehensive comparison between different techniques, as well as identifying the pros and cons of various evaluation metrics are also provided in this survey. 
Finally, we conclude by summarizing key implications, future research directions, and the challenges facing the research area.
\end{abstract}


\maketitle

\section{Introduction}\label{sec:Introduction}
Text classification -- the procedure of designating pre-defined labels for text -- is an essential and significant task in many Natural Language Processing (NLP) applications, such as sentiment analysis \cite{DBLP:conf/acl/TaiSM15, DBLP:conf/icml/ZhuSG15}, topic labeling \cite{DBLP:journals/apin/ChenGL20, DBLP:journals/isci/ChenGL19}, question answering \cite{DBLP:conf/acl/KalchbrennerGB14, DBLP:conf/emnlp/LiuQCWH15} and dialog act classification \cite{DBLP:conf/naacl/LeeD16}. 
In the era of information explosion, it is time-consuming and challenging to process and classify large amounts of text data manually. 
Besides, the accuracy of manual text classification can be easily influenced by human factors, such as fatigue and expertise. 
It is desirable to use machine learning methods to automate the text classification procedure to yield more reliable and less subjective results. 
Moreover, this can also help enhance information retrieval efficiency and alleviate the problem of information overload by locating the required information.

\begin{figure}[t]
    \centering
    \includegraphics[width=\linewidth]{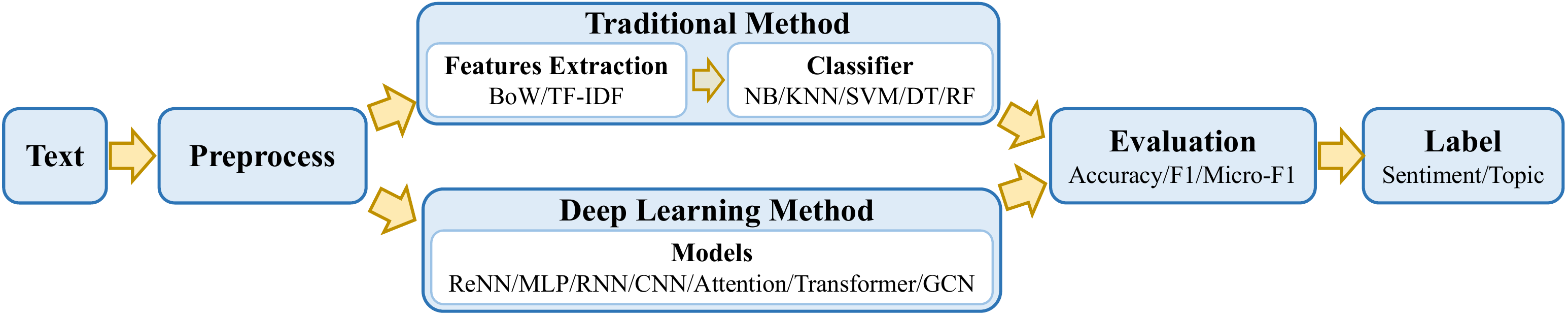}
    \caption{Flowchart of the text classification with classic methods in each module. It is crucial to extract essential features for traditional methods, but features can be extracted automatically by deep learning methods.}
    \label{fig:figure222}
    \label{img}
\end{figure}

Fig.~\ref{fig:figure222} illustrates a flowchart of the procedures involved in the text classification, under the light of traditional and deep analysis. Text data is different from numerical, image, or signal data. It requires NLP techniques to be processed carefully. The first important step is to preprocess text data for the model. Traditional models usually need to obtain good sample features by artificial methods and then classify them with classic machine learning algorithms. Therefore, the effectiveness of the method is largely restricted by feature extraction. However, different from traditional models, deep learning integrates feature engineering into the model fitting process by learning a set of nonlinear transformations that serve to map features directly to outputs.

From the 1960s until the 2010s, traditional text classification models dominated. 
Traditional methods mean statistics-based models, such as Naïve Bayes (NB) \cite{DBLP:journals/jacm/Maron61}, K-Nearest Neighbor (KNN) \cite{DBLP:journals/tit/CoverH67}, and Support Vector Machine (SVM) \cite{DBLP:conf/ecml/Joachims98}. 
Comparing with the earlier rule-based methods, this method has obvious advantages in accuracy and stability. 
However, these approaches still need to do feature engineering, which is time-consuming and costly. 
Besides, they usually disregard the natural sequential structure or contextual information in textual data, making it challenging to learn the semantic information of the words. 
Since the 2010s, text classification has gradually changed from traditional models to deep learning models. 
Compared with the methods based on traditional, deep learning methods avoid designing rules and features by humans and automatically provide semantically meaningful representations for text mining. 
Therefore, most of the text classification research works are based on Deep Neural Networks (DNNs) \cite{DBLP:conf/acl/AlyRB19}, which are data-driven approaches with high computational complexity. Few works focus on traditional models to settle the limitations of computation and data.

\subsection{Major Differences and Contributions}

There have been several works reviewing text classification and its subproblems recently. 
Two of them are reviews of text classification.
Kowsari et al. \cite{DBLP:journals/information/KowsariMHMBB19} surveyed different text feature extraction, dimensionality reduction methods, basic model structure for text classification, and evaluation methods. 
Minaee et al. \cite{DBLP:journals/corr/abs-2004-03705} reviewed recent deep learning based text classification methods, benchmark datasets, and evaluation metrics. 
Unlike existing text classification reviews, we conclude existing models from traditional models to deep learning with works of recent years.
Traditional models emphasize the feature extraction and classifier design. 
Once the text has well-designed characteristics, it can be quickly converged by training the classifier. 
DNNs can perform feature extraction automatically and learn well without domain knowledge. 
We then give the datasets and evaluation metrics for single-label and multi-label tasks and summarize future research challenges from data, models, and performance perspective. 
Moreover, we summarize various information in three tables, including the necessary information of classic deep learning models, primary information of main datasets, and a general benchmark of state-of-the-art methods under different applications. 
In summary, this study's main contributions are as follows:

\begin{itemize}
    \item We introduce the process and development of text classification and present comprehensive analysis and research on primary models -- from traditional to deep learning models -- according to their model structures. We summarize the necessary information of deep learning models in terms of basic model structures in Table~\ref{tab:BasicInformation}, including publishing years, methods, venues, applications, evaluation metrics, datasets and code links.
    \item We introduce the present datasets and give the formulation of main evaluation metrics with the comparison of metrics, including single-label and multi-label text classification tasks. We summarize the necessary information of primary datasets in Table~\ref{tab:datasets}, including the number of categories, average sentence length, the size of each dataset, related papers and data addresses.
    \item We summarize classification accuracy scores of models given in their articles, on benchmark datasets in Table~\ref{tab:Performance} and conclude the survey by discussing the main challenges facing the text classification and key implications stemming from this study.
\end{itemize}

\subsection{Organization of the Survey}

The rest of the survey is organized as follows. 
Section~\ref{Section 3} summarizes the existing models related to text classification, including traditional and deep learning models, including a summary table. 
Section~\ref{Section 4} introduces the primary datasets with a summary table and evaluation metrics on single-label and multi-label tasks. 
We then give quantitative results of the leading models in classic text classification datasets in Section~\ref{Section 5}. 
Finally, we summarize the main challenges for deep learning text classification in Section~\ref{Section 6} before concluding the article in Section~\ref{Section 7}.

\section{Text Classification Methods} \label{Section 3}

Text classification is referred to as extracting features from raw text data and predicting the categories of text data based on such features. 
Numerous models have been proposed in the past few decades for text classification. 
For traditional models, NB \cite{DBLP:journals/jacm/Maron61} is the first model used for the text classification task. 
Whereafter, generic classification models are proposed, such as KNN \cite{DBLP:journals/tit/CoverH67}, SVM~\cite{DBLP:conf/ecml/Joachims98}, and Random Forest (RF)~\cite{DBLP:journals/ml/Breiman01}, which are called classifiers, widely used for text classification. 
Recently, the eXtreme Gradient Boosting (XGBoost) \cite{DBLP:conf/kdd/ChenG16} and the Light Gradient Boosting Machine (LightGBM) \cite{DBLP:conf/nips/KeMFWCMYL17} have arguably the potential to provide excellent performance. 
For deep learning models, TextCNN \cite{DBLP:conf/emnlp/Kim14} has the highest number of references in these models, wherein a Convolutional Neural Network (CNN) \cite{albawi2017understanding} model has been introduced to solve the text classification problem for the first time. 
While not specifically designed for handling text classification tasks, the Bidirectional Encoder Representation from Transformers (BERT)~\cite{DBLP:conf/naacl/DevlinCLT19} has been widely employed when designing text classification models, considering its effectiveness on numerous text classification datasets.

\subsection{Traditional Models}

Traditional models accelerate text classification with improved accuracy and make the application scope of traditional expand.
The first thing is to preprocess the raw input text for training traditional models, which generally consists of word segmentation, data cleaning, and statistics.
Then, text representation aims to express preprocessed text in a form that is much easier for computers and minimizes information loss, such as Bag-Of-Words (BOW) \cite{zhang2010understanding}, N-gram \cite{cavnar1994n}, Term Frequency-Inverse Document Frequency (TF-IDF)~\cite{DBLP:reference/db/X09xxgr}, word2vec \cite{DBLP:journals/corr/abs-1301-3781} and Global Vectors for word representation (GloVe) ~\cite{DBLP:conf/emnlp/PenningtonSM14}. 
BOW means that all the words in the corpus are formed into a mapping array. According to the mapping array, a sentence can be represented as a vector. The $i$-th element in the vector represents the frequency of the $i$-th word in the mapping array of the sentence. The vector is the BOW of the sentence. At the core of the BOW is representing each text with a dictionary-sized vector. 
The individual value of the vector denotes the word frequency corresponding to its inherent position in the text. 
Compared to BOW, N-gram considers the information of adjacent words and builds a dictionary by considering the adjacent words. It is used to calculate the probability model of a sentence. The probability of a sentence is expressed as the joint probability of each word in the sentence. The probability of a sentence can be calculated by predicting the probability of the $N$-th word, given the sequence of the $(N-1)$-th words.
To simplify the calculation, the N-gram model adopts the Markov hypothesis \cite{cavnar1994n}. A word appears only concerning the words that preceded it. Therefore, the N-gram model performs a sliding window with size N. By counting and recording the occurrence frequency of all fragments, the probability of a sentence can be calculated using the frequency of relevant fragments in the record.
TF-IDF~\cite{DBLP:reference/db/X09xxgr} uses the word frequency and inverses the document frequency to model the text. 
TF is the word frequency of a word in a specific article, and IDF is the reciprocal of the proportion of the articles containing this word to the total number of articles in the corpus. TF-IDF is the multiplication of the two.
TF-IDF assesses the importance of a word to one document in a set of files or a corpus. The importance of a word increases proportionally with the number of times it appears in a document. However, it decreases inversely with its frequency in the corpus as a whole.
The word2vec \cite{DBLP:journals/corr/abs-1301-3781} employs local context information to obtain word vectors, as shown in Fig.~\ref{word2vec_Glove}. 
Word vector refers to a fixed-length real value vector specified as the word vector for any word in the corpus. The word2vec uses two essential models: CBOW and Skip-gram. The former is to predict the current word on the premise that the context of the current word is known. The latter is to predict the context when the current word is known.
The GloVe~\cite{DBLP:conf/emnlp/PenningtonSM14} -- with both the local context and global statistical features -- trains on the nonzero elements in a word-word co-occurrence matrix, as shown in Fig.~\ref{word2vec_Glove}. 
It enables word vectors to contain as much semantic and grammatical information as possible. The construction method of word vector is: firstly, the co-occurrence matrix of words is constructed based on the corpus, and then the word vector is learned based on the co-occurrence matrix and GloVe model.
Finally, the represented text is fed into the classifier according to selected features. 
Here, we discuss some representative classifiers in detail: 

\begin{figure*}[t]
    \centering
    \subfigure[CBOW.]{
 \includegraphics[width=5.7cm]{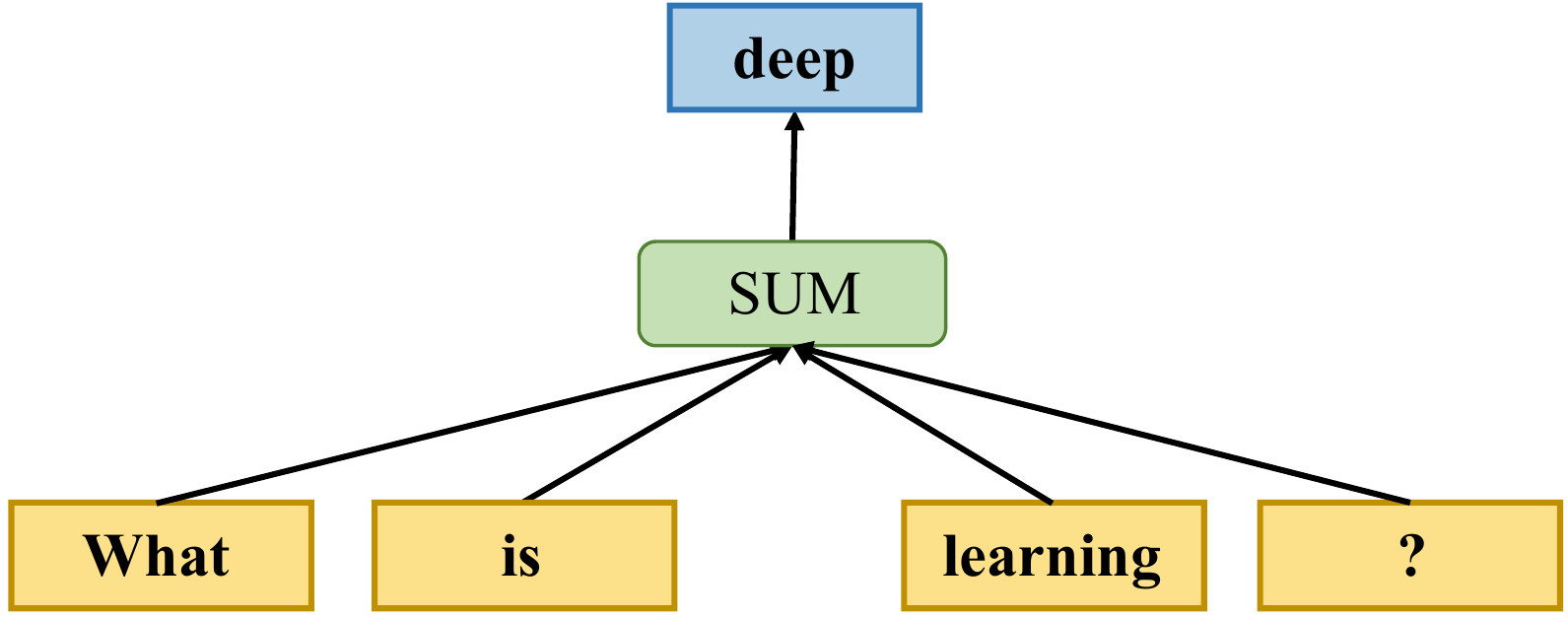}}
  \centering
 \subfigure[Skip-gram.]{
 \includegraphics[width=5.7cm]{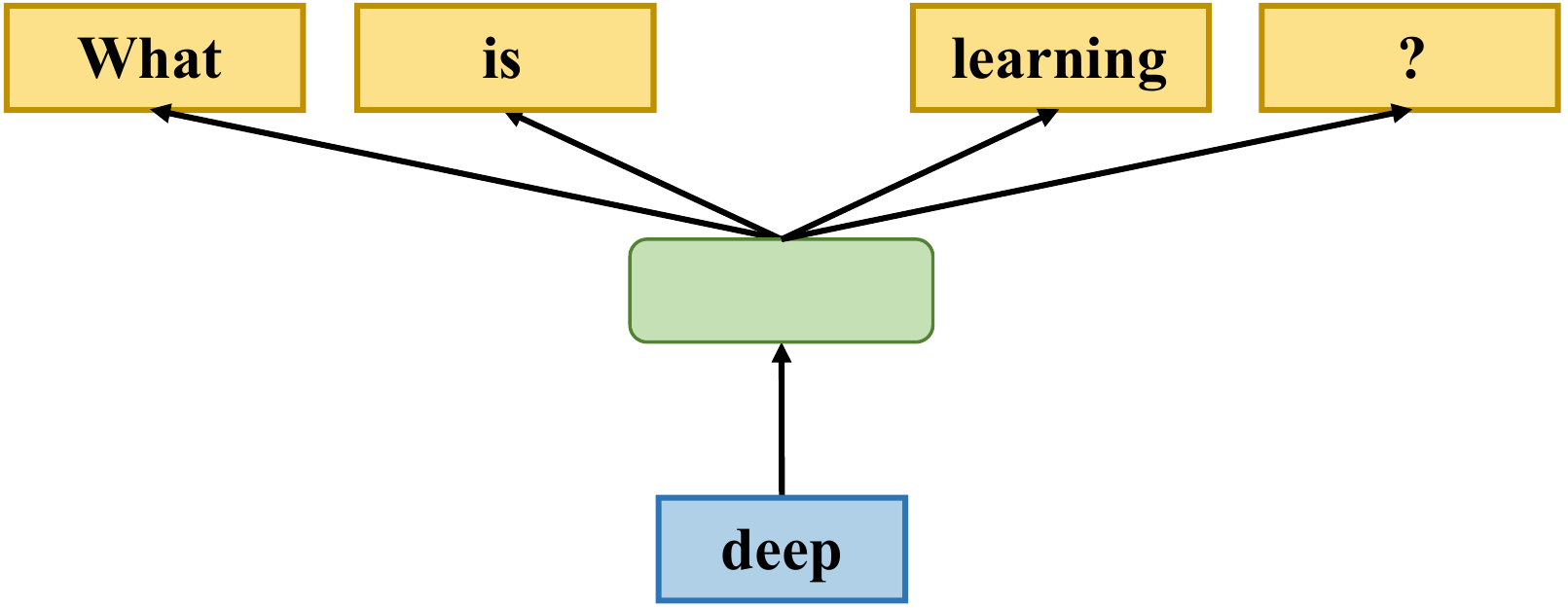}}
    \caption{The structure of word2vec, including CBOW and Skip-gram.}
    \label{dimensions}
\end{figure*}

\begin{figure}[!htbp]
    \centering
    \includegraphics[width=0.9\linewidth]{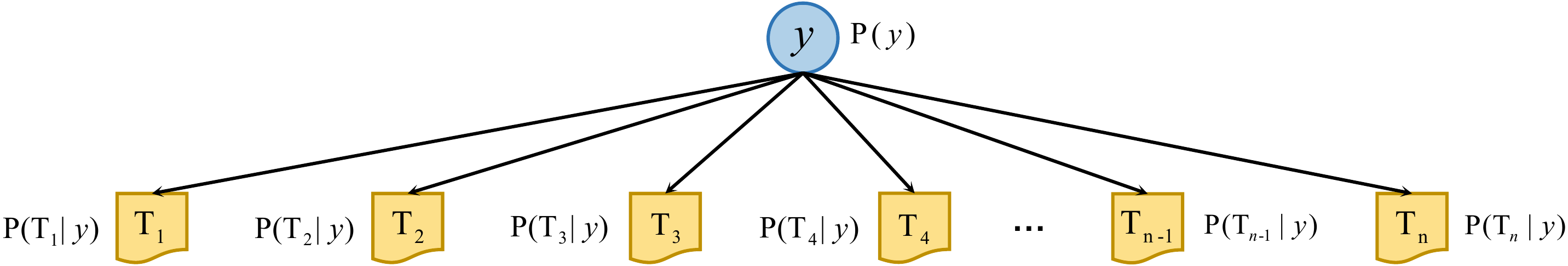}
    \caption{The structure of Naïve Bayes. }
    \label{figureNB_HMM}
\end{figure}

\subsubsection{PGM-based methods}

Probabilistic Graphical Models (PGMs) express the conditional dependencies among features in graphs, such as the Bayesian network \cite{DBLP:conf/kdd/ZhangZ10}. 
It is combinations of probability theory and graph theory.

Naïve Bayes (NB)~\cite{DBLP:journals/jacm/Maron61} is the simplest and most broadly used model based on applying Bayes' theorem. 
The NB algorithm has an independent assumption: when the target value has been given, the conditions between text $T=[T_{1}, T_{2},\cdots, T_{n}]$ are independent (see Fig.~\ref{figureNB_HMM}). 
The NB algorithm primarily uses the prior probability to calculate the posterior probability $\mathrm{P}\left(y \mid \mathrm{T}_{1}, T_{2}, \cdots, T_{n}\right)=\frac{p(y) \prod_{j=1}^{n} p\left(T_{j} \mid y\right)}{\prod_{j=1}^{\mathrm{n}} p\left(T_{j}\right)}$. 
Due to its simple structure, NB is broadly used for text classification tasks. 
Although the assumption that the features are independent is sometimes not actual, it substantially simplifies the calculation process and performs better.
To improve the performance on smaller categories, Schneider \cite{DBLP:conf/acl/Schneider04} proposes a feature selection score method through calculating KL-divergence \cite{10.5555/1146355} between the training set and corresponding categories for multinomial NB text classification.
Dai et al. \cite{DBLP:conf/aaai/DaiXYY07} propose a transfer learning method named Naive Bayes Transfer Classification (NBTC) to settle the different distribution between the training set and the target set. 
It uses the EM algorithm \cite{A1977Maximum} to obtain a locally optimal posterior hypothesis on the target set.
NB classifier is also used for fake news detection \cite{8100379}, and sentiment analysis \cite{inproceedings2017}, which can be seen as a text classification task.
Bernoulli NB, Gaussian NB and Multinomial NB are three popular approaches of NB text classification \cite{DBLP:journals/jis/Xu18}.
Multinomial NB performs slightly better than Bernoulli NB on few labeled dataset\cite{8776800}.
Bayesian NB classifier with Gaussian event model \cite{DBLP:journals/jis/Xu18} has been proven to be superior to NB with multinomial event model on 20 Newsgroups (20NG) \cite{datasets-for-single-label-textcategorization} and WebKB \cite{DBLP:conf/aaai/CravenFMMNS98} datasets.

\begin{figure}[!htbp]
    \centering
    \includegraphics[width=\linewidth]{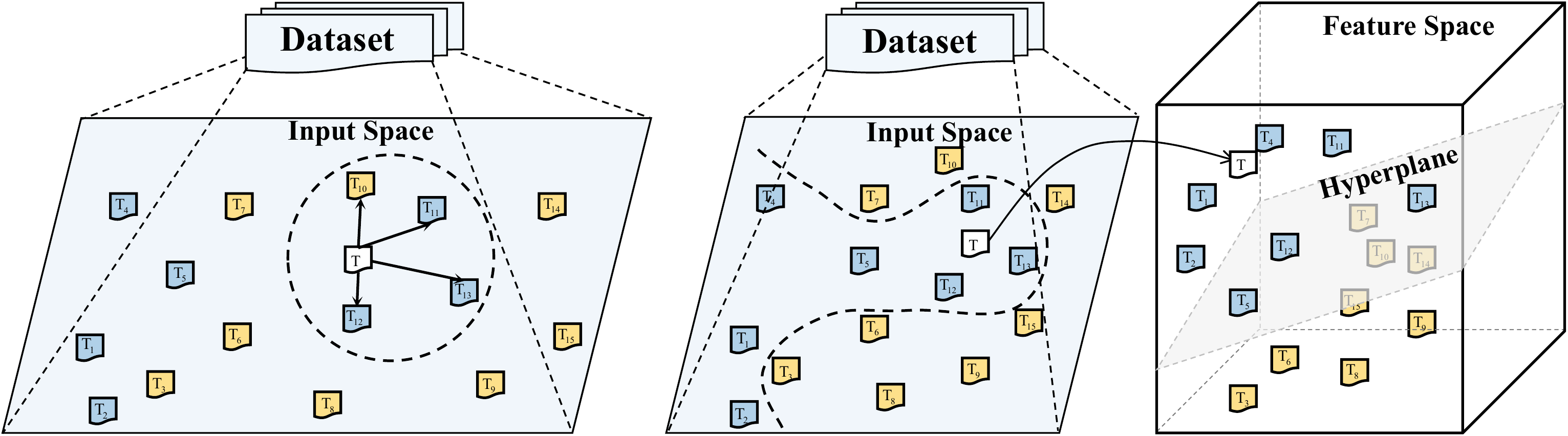}
    \caption{The structure of KNN where $k=4$ (left) and the structure of SVM (right). Each node represents a text and nodes with different contours represent different categories.}
    \label{figureKNN-SVM}
\end{figure}

\subsubsection{KNN-based Methods}

At the core of the K-Nearest Neighbors (KNN) algorithm \cite{DBLP:journals/tit/CoverH67} is to classify an unlabeled sample by finding the category with most samples on the $k$ nearest samples.
It is a simple classifier without building the model and can decrease complexity through the fasting process of getting $k$ nearest neighbors.
Fig.~\ref{figureKNN-SVM} showcases the structure of KNN. 
We can find $k$ training texts approaching a specific text to be classified through estimating the in-between distance. 
Hence, the text can be divided into the most common categories found in $k$ training set texts. 
The improvement of KNN algorithm mainly includes feature similarity \cite{7866706}, $K$ value \cite{10.1145/1039621.1039623} and index optimization \cite{Chen_2018}.
However, due to the positive correlation between model time/space complexity and the amount of data, the KNN algorithm takes an unusually long time on the large-scale datasets \cite{DBLP:journals/eswa/JiangPWK12}. 
To decrease the number of selected features, Soucy et al.~\cite{DBLP:conf/icdm/SoucyM01} propose a KNN algorithm without feature weighting. 
It manages to find relevant features, building the inter-dependencies of words by using a feature selection. 
When the data is extremely unevenly distributed, KNN tends to classify samples with more data. 
The Neighbor-Weighted K-Nearest Neighbor (NWKNN) \cite{DBLP:journals/eswa/Tan05} is proposed to improve classification performance on the unbalanced corpora.
It casts a significant weight for neighbors in a small category and a small weight for neighbors in a broad class.

\subsubsection{SVM-based Methods}

Cortes and Vapnik \cite{DBLP:journals/ml/CortesV95} propose Support Vector Machine (SVM) to tackle the binary classification of pattern recognition. 
Joachims~\cite{DBLP:conf/ecml/Joachims98}, for the first time, uses the SVM method for text classification representing each text as a vector. As illustrated in Fig.~\ref{figureKNN-SVM}, SVM-based approaches turn text classification tasks into multiple binary classification tasks. 
In this context, SVM constructs an optimal hyperplane in the one-dimensional input space or feature space, maximizing the distance between the hyperplane and the two categories of training sets, thereby achieving the best generalization ability. 
The goal is to make the distance of the category boundary along the direction perpendicular to the hyperplane is the largest. 
Equivalently, this will result in the lowest error rate of classification. 
Constructing an optimal hyperplane can be transformed into a quadratic programming problem to obtain a globally optimal solution.
Choosing the appropriate kernel function \cite{leslie2001spectrum} and feature selection \cite{taira1999feature} are of the utmost importance to ensure SVM can deal with nonlinear problems and become a robust nonlinear classifier. 
Furthermore, active learning \cite{li2013active} and adaptive learning \cite{peng2008svm} method are used for text classification to reduce the labeling effort based on the supervised learning algorithm SVM.
To analyze what the SVM algorithms learn and what tasks are suitable, Joachims \cite{DBLP:conf/sigir/Joachims01} proposes a theoretical learning model combining the statistical traits with the generalization performance of an SVM, analyzing the features and benefits using a quantitative approach. 
Transductive Support Vector Machine (TSVM) \cite{JOACHIMS1999Transductive} is proposed to lessen misclassifications of the particular test collections with a general decision function considering a specific test set.
It uses prior knowledge to establish a more suitable structure and study faster.

\begin{figure}[!htbp]
    \centering
    \includegraphics[width=\linewidth]{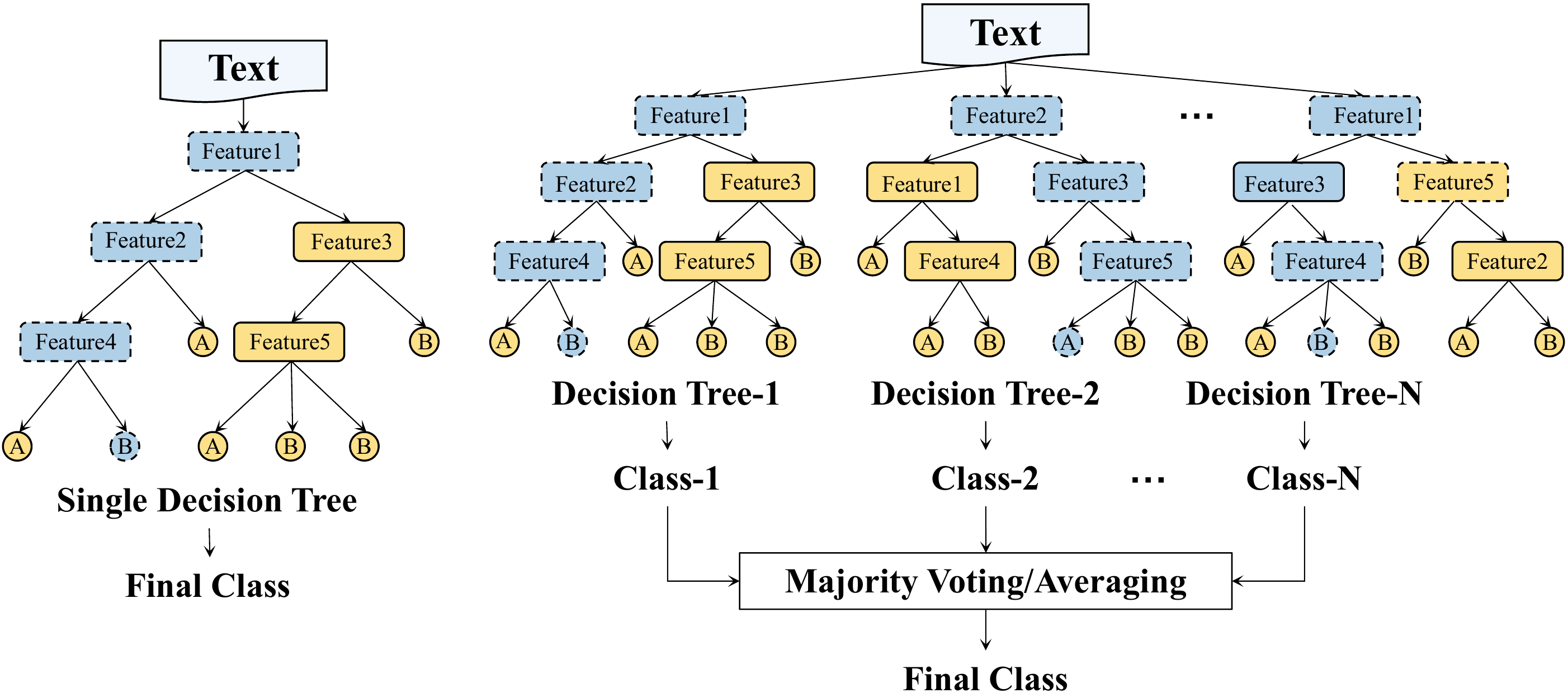}
    \caption{An example of DT (left) and the structure of RF (right). The nodes with the dotted outline represent the nodes of the decision route. It has five features to predict whether each text belongs to category A or B.}
    \label{figureDT-RF}
\end{figure}

\subsubsection{DT-based Methods}

Decision Trees (DT) \cite{DBLP:books/daglib/0087929} is a supervised tree structure learning method -- reflective of the idea of divide-and-conquer -- and is constructed recursively. 
It learns disjunctive expressions and has robustness for the text with noise. 
As shown in Fig.~\ref{figureDT-RF}, decision trees can be generally divided into two distinct stages: tree construction and tree pruning. 
It starts at the root node and tests the data samples (composed of instance sets, which have several attributes), and divides the dataset into diverse subsets according to different results. 
A subset of datasets constitutes a child node, and every leaf node in the decision tree represents a category. 
Constructing the decision tree is to determine the correlation between classes and attributes, further exploited to predict the record categories of unknown forthcoming types. 
The classification rules generated by the decision tree algorithm are straight-forward, and the pruning strategy \cite{DBLP:journals/datamine/RastogiS00} can also help reduce the influence of noise. 
Its limitation, however, mainly derives from inefficiency in coping with explosively increasing data size. 
More specifically, the Iterative Dichotomiser 3 (ID3) \cite{Ross1986Induction} algorithm uses information gain as the attribute selection criterion in the selection of each node -- It is used to select the attribute of each branch node, and then select the attribute having the maximum information gain value to become the discriminant attribute of the current node. 
Based on ID3, C4.5 \cite{10.5555/152181} learns to obtain a map from attributes to classes, which effectively classifies entities unknown to new categories. 
DT based algorithms usually need to train for each dataset, which is low efficiency \cite{kamber1997generalization}. Thus, Johnson et al. \cite{DBLP:journals/ibmsj/JohnsonOZG02} propose a DT-based symbolic rule system. 
The method represents each text as a vector calculated by the frequency of each word in the text, and induces rules from the training data. 
The learning rules are used for classifying the other data, being similar to the training data. 
Furthermore, to reduce the computational costs of DT algorithms, Fast Decision-Tree (FDT) \cite{DBLP:conf/icdm/VateekulK09} uses a two-pronged strategy: pre-selecting a feature set and training multiple DTs on different data subsets. 
Results from multiple DTs are combined through a data-fusion technique to resolve the cases of imbalanced classes.

\subsubsection{Integration-based Methods}

Integrated algorithms aim to aggregate the results of multiple algorithms for better performance and interpretation. 
Conventional integrated algorithms are bootstrap aggregation, such as RF \cite{DBLP:journals/ml/Breiman01}, boosting such as the Adaptive Boosting (AdaBoost) \cite{DBLP:conf/eurocolt/FreundS95}, and XGBoost \cite{DBLP:conf/kdd/ChenG16} and stacking. 
The bootstrap aggregation method trains multiple classifiers without strong dependencies and then aggregates their results.
For instance, RF \cite{DBLP:journals/ml/Breiman01} consists of multiple tree classifiers wherein all trees depend on the value of the random vector sampled independently (depicted in Fig.~\ref{figureDT-RF}). 
It is worth noting that each tree within the RF shares the same distribution. 
The generalization error of an RF relies on the strength of each tree and the relationship among trees, and will converge to a limit with the increment of tree number in the forest. 
In boosting based algorithms, all labeled data are trained with the same weight to initially obtain a weaker classifier \cite{DBLP:journals/ml/SchapireS99}. 
The weights of the data will then be adjusted according to the former result of the classifier. 
The training procedure will continue by repeating such steps until the termination condition is reached. 
Unlike bootstrap and boosting algorithms, stacking based algorithms break down the data into $n$ parts and use $n$ classifiers to calculate the input data in a cascade manner -- Result from upstream classifier will feed into the downstream classifier as input.
The training will terminate once a pre-defined iteration number is targeted. 
The integrated method can capture more features from multiple trees. 
However, it helps little for short text. 
Motivated by this, Bouaziz et al. ~\cite{DBLP:conf/dawak/BouazizDPPL14} combine data enrichment -- with semantics in RFs for short text classification -- to overcome the deficiency of sparseness and insufficiency of contextual information. 
In integrated algorithms, not all classifiers learn well. 
It is necessary to give different weights for each classifier. 
To differentiate contributions of trees in a forest, Islam et al. ~\cite{DBLP:conf/cikm/IslamLL0019} exploit the Semantics Aware Random Forest (SARF) classifier, choosing features similar to the features of the same class, for extracting features and producing the prediction values.

\textbf{Summary.}
The parameters of NB are more diminutive, less sensitive to missing data, and the algorithm is simple. However, it assumes that features are independent of each other. When the number of features is large, or the correlation between features is significant, the performance of NB decreases. SVM can solve high-dimensional and nonlinear problems. It has a high generalization ability, but it is sensitive to missing data. KNN mainly depends on the surrounding finite adjacent samples, rather than discriminating class domain to determine the category. Thus, for the dataset to be divided with more crossover or overlap of the class domain, it is more suitable than other methods. DT is easy to understand and interpret. Given an observed model, it is easy to deduce the corresponding logical expression according to the generated decision tree.
The traditional method is a type of machine learning. It learns from data, which are pre-defined features that are important to the performance of prediction values. 
However, feature engineering is tough work. 
Before training the classifier, we need to collect knowledge or experience to extract features from the original text.
The traditional methods train the initial classifier based on various textual features extracted from the raw text. 
Toward small datasets, traditional models usually present better performance than deep learning models under the limitation of computational complexity. 
Therefore, some researchers have studied the design of traditional models for specific domains with fewer data.

\subsection{Deep Learning Models}

The DNNs consist of artificial neural networks that simulate the human brain to automatically learn high-level features from data, getting better results than traditional models in speech recognition, image processing, and text understanding. 
Input datasets should be analyzed to classify the data, such as a single-label, multi-label, unsupervised, unbalanced dataset. 
According to the trait of the dataset, the input word vectors are sent into the DNN for training until the termination condition is reached. 
The performance of the training model is verified by the downstream task, such as sentiment classification, question answering, and event prediction. 
We show some DNNs over the years in Table \ref{tab:BasicInformation}, including designs that are different from the corresponding basic models, evaluation metrics, and experimental datasets.

\begin{table*}[!htbp]
\centering
\caption{Basic information based on different models. Trans: Transformer. Time: training time.}
\label{tab:BasicInformation}
\resizebox{\textwidth}{!}{
    \begin{tabular}{c|ccccccl} 
        \toprule
       \textbf{Model} & \textbf{Year} & \textbf{Method} & \textbf{Venue} & \textbf{Applications} & \textbf{Code Link} & \textbf{Metrics} & \textbf{Datasets} \\ 
       \midrule
        & 2011&RAE \cite{DBLP:conf/emnlp/SocherPHNM11}&EMNLP&SA, QA& \cite{Semi-Supervised-Recursive-Autoencoders-for-Predicting-Sentiment-Distributions} &Accuracy& MPQA, MR, EP \\ \cline{2-8}
        ReNN&2012&MV-RNN \cite{DBLP:conf/emnlp/SocherHMN12}&EMNLP&SA&\cite{MV_RNN}&Accuracy, F1& MR \\ \cline{2-8}
        &2013& RNTN \cite{DBLP:conf/emnlp/SocherPWCMNP13}&EMNLP&SA&\cite{DeepSentiment}&Accuracy& SST \\ \cline{2-8}
        &2014&DeepRNN \cite{DBLP:conf/nips/IrsoyC14}&NIPS&SA;QA&- &Accuracy&SST-1;SST-2\\ \hline

        MLP&2014&Paragraph-Vec \cite{DBLP:conf/icml/LeM14}&ICML&SA, QA&\cite{paragraph-vectors} &Error Rate& SST, IMDB \\ \cline{2-8}
        &2015&DAN \cite{DBLP:conf/acl/IyyerMBD15}&ACL&SA, QA&\cite{dan}  &Accuracy, Time&RT, SST, IMDB \\ 
        \hline
        
        & 2015&Tree-LSTM \cite{DBLP:conf/acl/TaiSM15}&ACL&SA& \cite{TreeLSTMSentiment} &Accuracy& SST-1, SST-2 \\ \cline{2-8}
         & 2015&S-LSTM \cite{DBLP:conf/icml/ZhuSG15}&ICML&SA&-& Accuracy& SST\\ \cline{2-8}
         & 2015  &  TextRCNN \cite{DBLP:conf/aaai/LaiXLZ15} &AAAI&SA, TL&\cite{rcnn-text-classification}& Macro-F1, etc.& 20NG, Fudan, ACL, SST-2 \\ \cline{2-8}
         & 2015&MT-LSTM \cite{DBLP:conf/emnlp/LiuQCWH15}&EMNLP&SA,QA&\cite{MT-LSTM} &Accuracy& SST-1, SST-2, QC, IMDB \\ \cline{2-8}
          & 2016&oh-2LSTMp \cite{DBLP:conf/icml/JohnsonZ16}&ICML&SA, TL&\cite{oh-2LSTMp} &Error Rate& IMDB, Elec, RCV1, 20NG \\ \cline{2-8}
         RNN& 2016&BLSTM-2DCNN \cite{DBLP:conf/coling/ZhouQZXBX16}&COLING&SA, QA, TL& \cite{NNForTextClassification} &Accuracy& SST-1, Subj, TREC, etc. \\ \cline{2-8}
         &2016&Multi-Task \cite{DBLP:conf/ijcai/LiuQH16}&IJCAI&SA&\cite{text_classification}  &Accuracy&SST-1, SST-2, Subj, IMDB \\ \cline{2-8}
         &2017  &   DeepMoji \cite{DBLP:conf/emnlp/FelboMSRL17}&EMNLP&SA&\cite{DeepMoji}    &Accuracy&SS-Twitter, SE1604, etc.  \\\cline{2-8}
         & 2017&TopicRNN \cite{DBLP:conf/iclr/Dieng0GP17}&ICML&SA&\cite{topic-rnn}  &Error Rate& IMDB \\ \cline{2-8}
         &2017&Miyato et al. \cite{DBLP:conf/iclr/MiyatoDG17}&ICLR&SA&\cite{adversarial_text}&Error Rate&IMDB, DBpedia, etc.\\ \cline{2-8}
         & 2018  &  RNN-Capsule \cite{DBLP:conf/www/WangSH0Z18}&TheWebConf&SA&\cite{Sentiment-Analysis-by-Capsules} &Accuracy&MR, SST-1, etc. \\ \cline{2-8}
         
         &2019&HM-DenseRNNs \cite{DBLP:conf/ijcai/ZhaoSY19}&IJCAI&SA, TL&\cite{HM-DenseRNNs}&Accuracy &IMDB, SST-5, AG\\ 
         \hline
         
        & 2014& TextCNN \cite{DBLP:conf/emnlp/Kim14} &EMNLP&SA, QA&\cite{CNN-for-Sentence-Classification-in-Keras}&Accuracy& MR, SST-2, Subj, etc. \\  \cline{2-8}
         & 2014&DCNN \cite{DBLP:conf/acl/KalchbrennerGB14} &ACL&SA, QA&\cite{ATS_Project}  &Accuracy&MR, TREC, Twitter\\ \cline{2-8} 
         & 2015   &  CharCNN \cite{DBLP:conf/nips/ZhangZL15}&NeurIPS&SA, QA, TL& \cite{CharCNN} &Error Rate&AG, Yelp P, DBPedia, etc.\\ \cline{2-8}
         &2016  &   SeqTextRCNN \cite{DBLP:conf/naacl/LeeD16} &NAACL&Dialog act&\cite{short-text-classification} & Accuracy &DSTC 4, MRDA, SwDA\\\cline{2-8}
        &2017&XML-CNN \cite{DBLP:conf/sigir/LiuCWY17}&SIGIR&NC, TL, SA&\cite{XML-CNN}&NDCG@K, etc. &EUR-Lex, Wiki-30K, etc.\\\cline{2-8}
         CNN& 2017   &  DPCNN \cite{DBLP:conf/acl/JohnsonZ17} &ACL&SA, TL&\cite{DPCNN}  &Error Rate&AG, DBPedia, Yelp.P, etc. \\ \cline{2-8} 
         & 2017&KPCNN \cite{DBLP:conf/ijcai/WangWZY17} &IJCAI&SA, QA, TL&- &Accuracy&Twitter, AG, Bing, etc. \\ \cline{2-8}
         & 2018  &  TextCapsule \cite{DBLP:conf/emnlp/YangZYLZZ18} &EMNLP&SA, QA, TL& \cite{capsule_text_classification}  &Accuracy&Subj, TREC, Reuters, etc. \\\cline{2-8} 
        & 2018  &   HFT-CNN \cite{DBLP:conf/emnlp/ShimuraLF18} &EMNLP&TL&\cite{HFT-CNN}  &Micro-F1, etc.&  RCV1, Amazon670K \\\cline{2-8} 
        & 2019  &   CCRCNN \cite{DBLP:conf/aaai/XuC19} &AAAI&TL& -  &Accuracy&  TREC, MR, AG \\\cline{2-8} 
        &2020&Bao et al. \cite{DBLP:conf/iclr/BaoWCB20}&ICLR&TL&\cite{Distributional-Signatures}&Accuracy&20NG, Reuters-2157, etc.\\         \hline 
         
         &   2016  &  HAN \cite{DBLP:conf/naacl/YangYDHSH16} &NAACL&SA, TL&\cite{textClassifier} &Accuracy&Yelp.F, YahooA, etc.\\ \cline{2-8}
        & 2016&BI-Attention \cite{DBLP:conf/emnlp/ZhouWX16}&NAACL&SA&-&Accuracy&NLP\&CC 2013 \cite{tcci.ccf.org.cn} \\  \cline{2-8}
        & 2016&LSTMN \cite{DBLP:conf/emnlp/0001DL16}&EMNLP&SA&\cite{Abstractive-Summarization}&Accuracy& SST-1\\ \cline{2-8}
        &2017&Lin et al. \cite{DBLP:conf/iclr/LinFSYXZB17}&ICLR&SA&\cite{Structured-Self-Attention}&Accuracy&Yelp, SNLI Age\\\cline{2-8}
        &  2018   &  SGM \cite{DBLP:conf/coling/YangSLMWW18} &COLING&TL&\cite{SGM}  &HL, Micro-F1&RCV1-V2, AAPD \\\cline{2-8}
         &2018&ELMo \cite{DBLP:conf/naacl/PetersNIGCLZ18} &NAACL&SA, QA, NLI&\cite{flair}&Accuracy&SQuAD, SNLI, SST-5\\ \cline{2-8}
         Attention&2018&BiBloSA \cite{DBLP:conf/iclr/ShenZL0Z18}&ICLR&SA&\cite{BiBloSA}&Accuracy, Time&CR, MPQA, SUBJ, etc.\\ \cline{2-8} 
         
          & 2019   &  AttentionXML \cite{DBLP:conf/nips/YouZWDMZ19}&NeurIPS&TL &\cite{AttentionXML} &P@k, N@k, etc. &EUR-Lex, etc. \\\cline{2-8}
         &2019&HAPN \cite{DBLP:conf/emnlp/SunSZL19}&EMNLP&RC&-&Accuracy&FewRel, CSID \\ \cline{2-8} 
         &2019&Proto-HATT \cite{DBLP:conf/aaai/GaoH0S19} &AAAI&RC&\cite{HATT-Proto}&Accuracy&FewRel\\ \cline{2-8} 
         &2019&STCKA \cite{DBLP:conf/aaai/ChenHLXJ19}&AAAI&SA, TL&\cite{STCKA} &Accuracy&Weibo, Product Review, etc. \\ \cline{2-8}
         &2020&HyperGAT \cite{DBLP:conf/emnlp/DingWLLL20}&EMNLP&TL, NC&\cite{HyperGAT} &Accuracy&20NG, Ohsumed, MR, etc. \\ \cline{2-8}
         &2020& MSMSA \cite{DBLP:conf/aaai/GuoQLXZ20}&AAAI&ST, QA, NLI&- &Accuracy, F1&IMDB, MR, SST, SNLI, etc. \\ \cline{2-8}
         &2020&Choi \cite{DBLP:conf/emnlp/ChoiPYH20} &EMNLP&SA, TL&-&Accuracy&SST2, IMDB, 20NG \\   \hline
        &2019 &  BERT \cite{DBLP:conf/naacl/DevlinCLT19}&NAACL&SA, QA&\cite{bert} &Accuracy&SST-2, QQP, QNLI, CoLA\\  \cline{2-8}
        & 2019  &   BERT-BASE \cite{DBLP:conf/acl/ChalkidisFMA19}&ACL&TL &\cite{lmtc-eurlex57k}&P@K, R@K, etc.&EUR-LEX   \\  \cline{2-8} 
        &2019&Sun et al. \cite{DBLP:conf/cncl/SunQXH19}&CCL&SA, QA, TL&\cite{How_Fin} &Error Rate&TREC, DBPedia, etc.\\ \cline{2-8} 
        & 2019&XLNet \cite{DBLP:conf/nips/YangDYCSL19}&NeurIPS&SA, QA, NC&\cite{xlnet}&EM, F1, etc. &Yelp-2, AG, MNLI, etc.\\ \cline{2-8} 
        &2019&RoBERTa \cite{DBLP:journals/corr/abs-1907-11692}&arXiv&SA, QA&\cite{RoBERTa}&F1, Accuracy&SQuAD, MNLI-m, SST-2\\ \cline{2-8} 
        Trans&2020&GAN-BERT \cite{DBLP:conf/acl/CroceCB20}&ACL&SA, NLI&\cite{GAN-BERT}&F1, Accuracy&SST-5, MNLI\\ \cline{2-8}
        &2020&BAE \cite{DBLP:conf/emnlp/GargR20}&EMNLP&SA, QA&\cite{BAE}&Accuracy&Amazon, Yelp, MR, MPQA\\ \cline{2-8}
        
        &2020&ALBERT \cite{DBLP:conf/iclr/LanCGGSS20}&ICLR&SA, QA&\cite{ALBERT}&F1, Accuracy& SST, MNLI, SQuAD\\ \cline{2-8}
        
         &2020&TG-Transformer \cite{DBLP:conf/emnlp/ZhangZ20}&EMNLP&SA, TL&-&Accuracy, Time& R8, R52, Ohsumed, etc.\\ \cline{2-8}
         &2020&X-Transformer \cite{DBLP:conf/kdd/ChangYZYD20}&KDD&SA, TL&\cite{X-Transformer}&P@K, R@K& Eurlex-4K, Wiki10-31K, etc.\\ \cline{2-8}
        
        &2021& LightXML \cite{DBLP:journals/corr/abs-2101-03305}&arXiv&TL, ML, NLI&\cite{LightXML}&P@K, Time & AmazonCat-13K, etc.\\
        \hline 
        &2018&DGCNN \cite{DBLP:conf/www/PengLHLBWS018} &TheWebConf&TL &\cite{DeepGraphCNNforTexts} &Macro-F1, etc. &RCV1, NYTimes\\ \cline{2-8}
        & 2019   &TextGCN \cite{DBLP:conf/aaai/YaoM019}&AAAI&SA, TL &\cite{text_gcn} &Accuracy&20NG, Ohsumed, R52, etc. \\\cline{2-8}
        &2019&SGC\cite{DBLP:conf/icml/WuSZFYW19} &ICML&NC, TL, SA & \cite{SGC}  &Accuracy, Time&20NG, R8, Ohsumed, etc.\\\cline{2-8}
        GNN&2019&Huang et al. \cite{DBLP:conf/emnlp/HuangMLZW19}&EMNLP&NC, TL&\cite{TextLevelGNN} &Accuracy&R8, R52, Ohsumed
        \\\cline{2-8}
        &2019&Peng et al. \cite{peng2019hierarchical} &arXiv&NC, TL& - &Micro-F1, etc. &RCV1, EUR-Lex, etc.\\ 
        \cline{2-8}
        &2020& TextING \cite{DBLP:conf/acl/ZhangYCWWW20} &ACL&SA, NC, TL& \cite{TextING} &Accuracy &MR, R8, R52, Ohsumed\\ 
        \cline{2-8}
        &2020& TensorGCN \cite{DBLP:conf/aaai/LiuYZWL20} &AAAI&SA, NC, TL& \cite{TensorGCN} &Accuracy &20NG, R8, R52, Ohsumed, MR\\ 
        \cline{2-8}
        &2020  &  MAGNET \cite{DBLP:conf/icaart/PalSS20} &ICAART&TL&\cite{MAGnet}  &Micro-F1, HL& Reuters, RCV1-V2, etc.   \\
        \hline 
        &2017&Miyato et al. \cite{DBLP:conf/iclr/MiyatoDG17}&ICLR&SA, NC&\cite{Miyato}&Error Rate&IMDB, RCV1, et al.\\ \cline{2-8} 
        Others&2018&TMN \cite{DBLP:conf/emnlp/ZengLSGLK18} &EMNLP&TL&-&Accuracy, F1&Snippets, Twitter, et al.\\ \cline{2-8} 
         &2019&Zhang et al. \cite{DBLP:conf/naacl/ZhangLG19}&NAACL&TL, NC&\cite{KG4ZeroShotText} &Accuracy&DBpedia, 20NG. \\   
         \bottomrule
    \end{tabular} }
\end{table*}

Numerous deep learning models have been proposed in the past few decades for text classification, as shown in Table~\ref{tab:BasicInformation}. 
We tabulate primary information -- including publication years, venues, applications, code links, evaluation metrics, and experiment datasets -- of main deep learning models for text classification. 
The applications in this table include Sentiment Analysis (SA), Topic Labeling (TL), News Classification (NC), Question Answering (QA), Dialog Act Classification (DAC), Natural Language Inference (NLI) and Relation Classification (RC). 
The multilayer perceptron \cite{4809024} and the recursive neural network \cite{DBLP:journals/csur/PouyanfarSYTTRS19} are the first two deep learning approaches used for the text classification task, which improve performance compared with traditional models. 
Then, CNNs, Recurrent Neural Networks (RNNs), and attention mechanisms are used for text classification \cite{DBLP:conf/emnlp/YangZYLZZ18, DBLP:conf/aaai/QinCLN020, DBLP:conf/naacl/DengPHLY21}. 
Many researchers advance text classification performance for different tasks by improving CNN, RNN, and attention, or model fusion and multi-task methods. 
The appearance of BERT \cite{DBLP:conf/naacl/DevlinCLT19}, which can generate contextualized word vectors, is a significant turning point in the development of text classification and other NLP technologies.
Many researchers \cite{DBLP:conf/aaai/JinJZS20, DBLP:conf/acl/CroceCB20} have studied text classification models based on BERT, which achieves better performance than the above models in multiple NLP tasks, including text classification. 
Besides, some researchers study text classification technology based on Graph Neural Network (GNN) \cite{DBLP:conf/aaai/YaoM019, lichen2021ijcai} to capture structural information in the text, which cannot be replaced by other methods. 
Here, we classify DNNs by structure and discuss some representative models in detail:

\begin{figure}[!htbp]
    \centering
    \includegraphics[width=\linewidth]{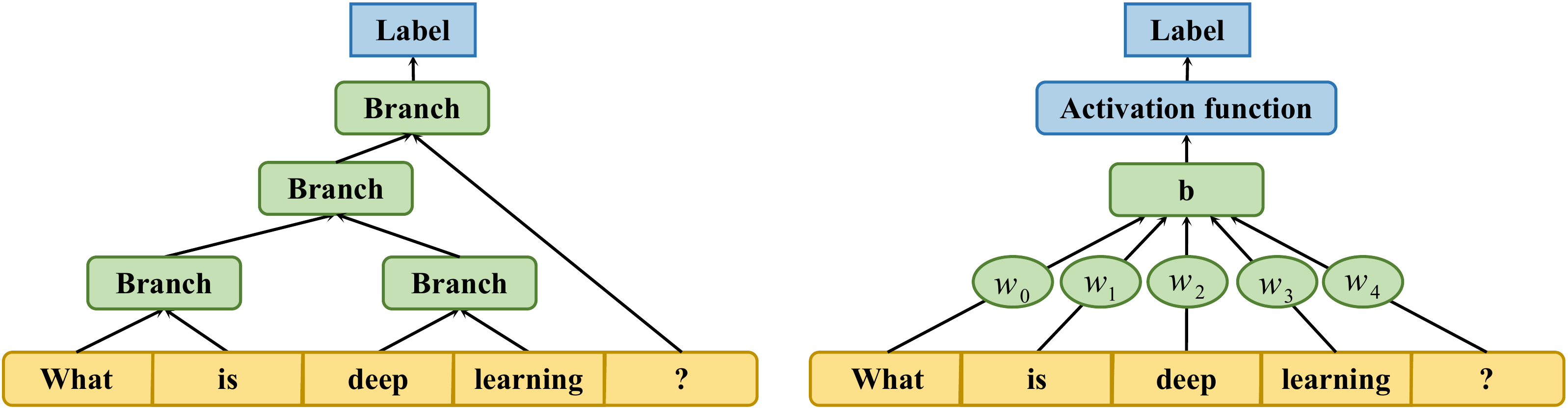}
    \caption{The architecture of ReNN (left) and the architecture of MLP (right).}
    \label{figureReNN-MLP}
\end{figure}

\subsubsection{ReNN-based Methods}

Traditional models cost lots of time on design features for each task. 
Furthermore, in the case of deep learning, the meaning of "word vectors" is different: each input word is associated with a fixed-length vector whose values are either drawn at random or derived from a previous traditional process, thus forming a matrix $L$ called word embedding matrix which represents the vocabulary words in a small latent semantic space, of generally 50 to 300 dimensions.
The Recursive Neural Network (ReNN) \cite{DBLP:journals/csur/PouyanfarSYTTRS19} can automatically learn the semantics of text recursively and the syntax tree structure without feature design, as shown in Fig.~\ref{figureReNN-MLP}. 
We give an example of ReNN based models. 
First, each word of input text is taken as the leaf node of the model structure. 
Then all nodes are combined into parent nodes using a weight matrix. 
The weight matrix is shared across the whole model. 
Each parent node has the same dimension with all leaf nodes.
Finally, all nodes are recursively aggregated into a parent node to represent the input text to predict the label.

ReNN-based models improve performance compared with traditional models and save on labor costs due to excluding feature designs used for different text classification tasks.
The Recursive AutoEncoder (RAE) \cite{DBLP:conf/emnlp/SocherPHNM11} is used to predict the distribution of sentiment labels for each input sentence and learn the representations of multi-word phrases. 
To learn compositional vector representations for each input text, the Matrix-Vector Recursive Neural Network (MV-RNN) \cite{DBLP:conf/emnlp/SocherHMN12} introduces a ReNN model to learn the representation of phrases and sentences. 
It allows that the length and type of input texts are inconsistent. 
MV-RNN allocates a matrix and a vector for each node on the constructed parse tree. 
Furthermore, the Recursive Neural Tensor Network (RNTN) \cite{DBLP:conf/emnlp/SocherPWCMNP13} is proposed with a tree structure to capture the semantics of sentences. 
It inputs phrases with different length and represents the phrases by parse trees and word vectors. 
The vectors of higher nodes on the parse tree are estimated by the equal tensor-based composition function.
For RNTN, the time complexity of building the textual tree is high, and expressing the relationship between documents is complicated within a tree structure.
The performance is usually improved, with the depth being increased for DNNs.
Therefore, Irsoy et al. \cite{DBLP:conf/nips/IrsoyC14} propose a Deep Recursive Neural Network (DeepReNN), which stacks multiple recursive layers.
It is built by binary parse trees and learns distinct perspectives of compositionality in language.

\subsubsection{MLP-based Methods}

A MultiLayer Perceptron (MLP) \cite{4809024}, sometimes colloquially called "vanilla" neural network, is a simple neural network structure that is used for capturing features automatically. 
As shown in Fig.~\ref{figureReNN-MLP}, we show a three-layer MLP model. 
It contains an input layer, a hidden layer with an activation function in all nodes, and an output layer. 
Each node connects with a certain weight $w_{i}$. 
It treats each input text as a bag of words and achieves high performance on many text classification benchmarks comparing with traditional models.

There are some MLP-based methods proposed by some research groups for text classification tasks. 
The Paragraph Vector (Paragraph-Vec) \cite{DBLP:conf/icml/LeM14} is the most popular and widely used method, which is similar to the Continuous Bag-Of-Words (CBOW) \cite{DBLP:journals/corr/abs-1301-3781}. 
It gets fixed-length feature representations of texts with various input lengths by employing unsupervised algorithms. 
Comparing with CBOW, it adds a paragraph token mapped to the paragraph vector by a matrix. 
The model predicts the fourth word by the connection or average of this vector to the three contexts of the word. 
Paragraph vectors can be used as a memory for paragraph themes and are used as a paragraph function and inserted into the prediction classifier.

\begin{figure}[!htbp]
    \centering
    \includegraphics[width=\linewidth]{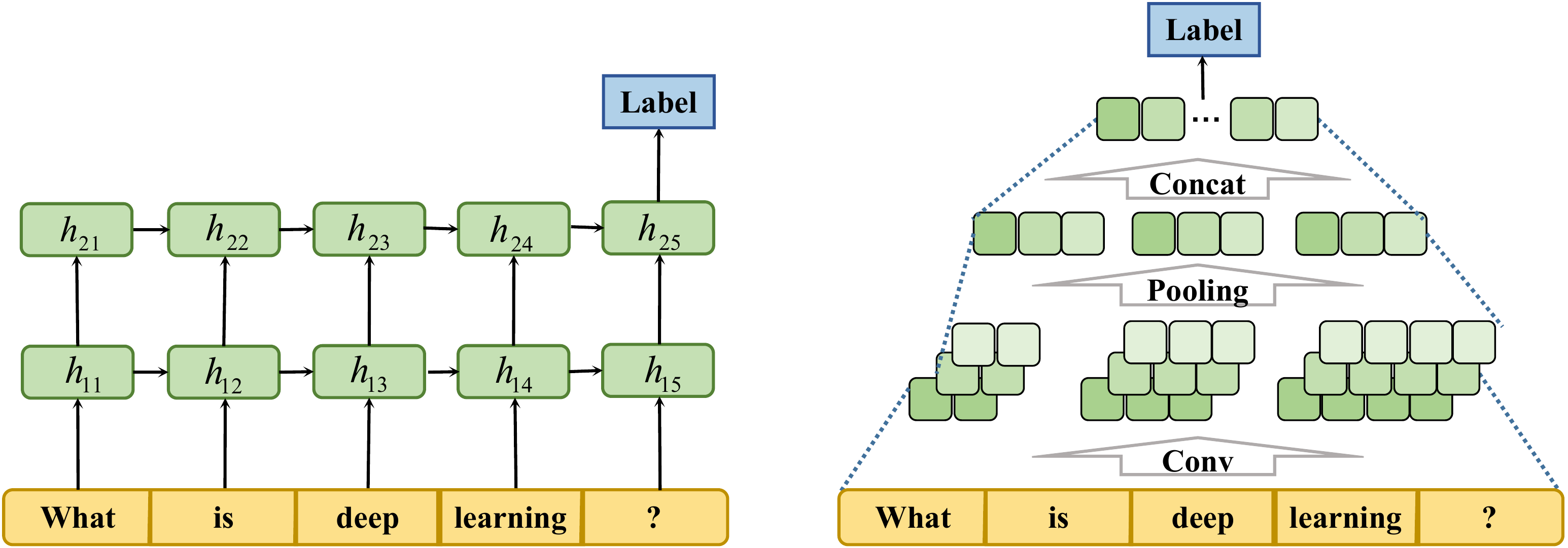}
    \caption{The RNN based model (left) and the CNN based model (right).}
    \label{figureCNN-RNN}
\end{figure}

\subsubsection{RNN-based Methods}

The Recurrent Neural Network (RNN) \cite{DBLP:journals/csur/PouyanfarSYTTRS19} is broadly used for capturing long-range dependency through recurrent computation. 
The RNN language model learns historical information, considering the location information among all words suitable for text classification tasks.
We show an RNN model for text classification with a simple sample, as shown in Fig.~\ref{figureCNN-RNN}. 
Firstly, each input word is represented by a specific vector using a word embedding technology. 
Then, the embedding word vectors are fed into RNN cells one by one. 
The output of RNN cells are the same dimension with the input vector and are fed into the next hidden layer. 
The RNN shares parameters across different parts of the model and has the same weights of each input word. 
Finally, the label of input text can be predicted by the last output of the hidden layer.

To diminish the time complexity of the model and capture contextual information, Liu et al. \cite{DBLP:conf/ijcai/LiuQH16} introduce a model for catching the semantics of long texts. 
It is a biased model that parsed the text one by one, making the following inputs profit over the former and decreasing the semantic efficiency of capturing the whole text.
For modeling topic labeling tasks with long input sequences, TopicRNN \cite{DBLP:conf/iclr/Dieng0GP17} is proposed. 
It captures the dependencies of words in a document via latent topics and uses RNNs to capture local dependencies and latent topic models for capturing global semantic dependencies. 
Virtual Adversarial Training (VAT) \cite{DBLP:journals/corr/MiyatoMKI17} is a useful regularization method applicable to semi-supervised learning tasks. 
Miyato et al. \cite{DBLP:conf/iclr/MiyatoDG17} apply adversarial and virtual adversarial training text and employ the perturbation into word embedding rather than the original input text. 
The model improves the quality of the word embedding and is not easy to overfit during training.
Capsule network \cite{10.1007/978-3-642-21735-7_6} captures the relationships between features using dynamic routing between capsules comprised of a group of neurons in a layer. 
Wang et al. \cite{DBLP:conf/www/WangSH0Z18} propose an RNN-Capsule model with a simple capsule structure for the sentiment classification task.

In the backpropagation process of RNN, the weights are adjusted by gradients, calculated by continuous multiplications of derivatives. 
If the derivatives are extremely small, it may cause a gradient vanishing problem by continuous multiplications. 
Long Short-Term Memory (LSTM) \cite{Hochreiter1997Long}, the improvement of RNN, effectively alleviates the gradient vanishing problem. 
It is composed of a cell to remember values on arbitrary time intervals and three gate structures to control information flow.
The gate structures include input gates, forget gates, and output gates. 
The LSTM classification method can better capture the connection among context feature words, and use the forgotten gate structure to filter useless information, which is conducive to improving the total capturing ability of the classifier. 
Tree-LSTM \cite{DBLP:conf/acl/TaiSM15} extends the sequence of LSTM models to the tree structure. 
The whole subtree with little influence on the result can be forgotten through the LSTM forgetting gate mechanism for the Tree-LSTM model.

Natural Language Inference (NLI) \cite{DBLP:conf/emnlp/BowmanAPM15} predicts whether one text's meaning can be deduced from another by measuring the semantic similarity between each pair of sentences.
To consider other granular matchings and matchings in the reverse direction, 
Wang et al. \cite{DBLP:conf/ijcai/WangHF17} propose a model for the NLI task named Bilateral Multi-Perspective Matching (BiMPM). 
It encodes input sentences by the BiLSTM encoder. 
Then, the encoded sentences are matched in two directions. 
The results are aggregated in a fixed-length matching vector by another BiLSTM layer. 
Finally, the result is evaluated by a fully connected layer.

\subsubsection{CNN-based Methods}

Convolutional Neural Networks (CNNs) \cite{albawi2017understanding} are proposed for image classification with convolving filters that can extract features of pictures. 
Unlike RNN, CNN can simultaneously apply convolutions defined by different kernels to multiple chunks of a sequence. 
Therefore, CNNs are used for many NLP tasks, including text classification. 
For text classification, the text requires being represented as a vector similar to the image representation, and text features can be filtered from multiple angles, as shown in Fig. \ref{figureCNN-RNN}. 
Firstly, the word vectors of the input text are spliced into a matrix. 
The matrix is then fed into the convolutional layer, which contains several filters with different dimensions. 
Finally, the result of the convolutional layer goes through the pooling layer and concatenates the pooling result to obtain the final vector representation of the text. The category is predicted by the final vector.

To try using CNN for the text classification task, an unbiased model of convolutional neural networks is introduced by Kim, called TextCNN \cite{DBLP:conf/emnlp/Kim14}. 
It can better determine discriminative phrases in the max-pooling layer with one layer of convolution and learn hyperparameters except for word vectors by keeping word vectors static. 
Training only on labeled data is not enough for data-driven deep models.
Therefore, some researchers consider utilizing unlabeled data. 
Johnson et al. \cite{DBLP:conf/nips/JohnsonZ15} propose a CNN model based on two-view semi-supervised learning for text classification, which first uses unlabeled data to train the embedding of text regions and then labeled data. 
DNNs usually have better performance, but it increases the computational complexity. 
Motivated by this, a Deep Pyramid Convolutional Neural Network (DPCNN) \cite{DBLP:conf/acl/JohnsonZ17} is proposed, with a little more computational accuracy, increasing by raising the network depth. 
The DPCNN is more specific than Residual Network (ResNet) \cite{DBLP:conf/eccv/HeZRS16}, as all the shortcuts are exactly simple identity mappings without any complication for dimension matching. 

According to the minimum embedding unit of text, embedding methods are divided into character-level, word-level, and sentence-level embedding. 
Character-level embeddings can settle Out-Of-Vocabulary (OOV) \cite{bazzi2002modelling} words. 
Word-level embeddings learn the syntax and semantics of the words. 
Moreover, sentence-level embedding can capture relationships among sentences. 
Motivated by these, Nguyen et al. \cite{DBLP:conf/pacling/NguyenN17} propose a deep learning method based on a dictionary, increasing information for word-level embeddings through constructing semantic rules and deep CNN for character-level embeddings. 
Adams et al. \cite{DBLP:journals/tgis/AdamsM18} propose a character-level CNN model, called MGTC, to classify multi-lingual texts written. 
TransCap \cite{DBLP:conf/acl/ChenQ19} is proposed to encapsulate the sentence-level semantic representations into semantic capsules and transfer document-level knowledge.

RNN based models capture the sequential information to learn the dependency among input words, and CNN based models extract the relevant features from the convolution kernels. 
Thus some works study the fusion of the two methods. BLSTM-2DCNN \cite{DBLP:conf/coling/ZhouQZXBX16} integrates a Bidirectional LSTM (BiLSTM) with two-dimensional max pooling.
It uses a 2D convolution to sample more meaningful information of the matrix and understands the context better through BiLSTM.
Moreover, Xue et al. \cite{DBLP:conf/ijcnlp/XueZLW17} propose MTNA, a combination of BiLSTM and CNN layers, to solve aspect category classification and aspect term extraction tasks.

\begin{figure}[!htbp]
    \centering
    \includegraphics[width=0.75\linewidth]{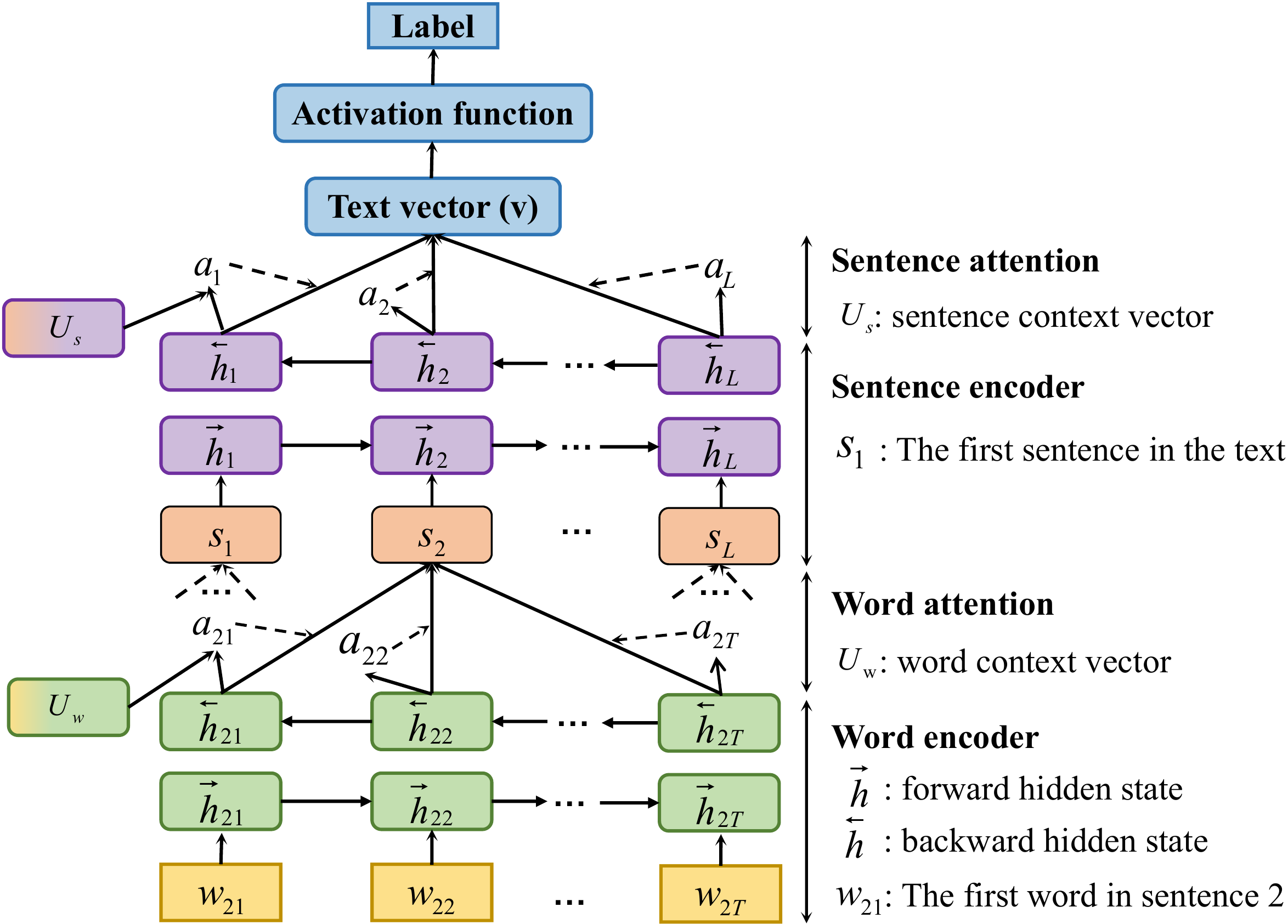}
    \caption{The architecture of hierarchical attention network (HAN) \cite{DBLP:conf/naacl/YangYDHSH16}.}
    \label{figureHAN}
\end{figure}  

\subsubsection{Attention-based Methods}

CNN and RNN provide excellent results on tasks related to text classification. 
However, these models are not intuitive enough for poor interpretability, especially in classification errors, which cannot be explained due to the non-readability of hidden data.
The attention-based methods are successfully used in the text classification. 
Bahdanau et al. \cite{DBLP:journals/corr/BahdanauCB14} first propose an attention mechanism that can be used in machine translation. 
Motivated by this, Yang et al. \cite{DBLP:conf/naacl/YangYDHSH16} introduce the Hierarchical Attention Network (HAN) to gain better visualization by employing the extremely informational components of a text, as shown in Fig.~\ref{figureHAN}. 
HAN includes two encoders and two levels of attention layers. 
The attention mechanism lets the model pay different attention to specific inputs. 
It aggregates essential words into sentence vectors firstly and then aggregates vital sentence vectors into text vectors. 
It can learn how much contribution of each word and sentence for the classification judgment, which is beneficial for applications and analysis through the two levels of attention.

The attention mechanism can improve the performance with interpretability for text classification, which makes it popular. 
There are some other works based on attention. 
LSTMN \cite{DBLP:conf/emnlp/0001DL16} is proposed to process text step by step from left to right and does superficial reasoning through memory and attention. 
BI-Attention \cite{DBLP:conf/emnlp/ZhouWX16} is designed for cross-lingual text classification to catch bilingual long-distance dependencies. 
Hu et al. \cite{DBLP:conf/coling/HuLT0S18} propose an attention mechanism based on category attributes for solving the imbalance of the number of various charges which contain few-shot charges. 
HAPN \cite{DBLP:conf/emnlp/SunSZL19} is presented for few-shot text classification.

\begin{figure}[!htbp]
    \centering
    \includegraphics[width=0.75\linewidth]{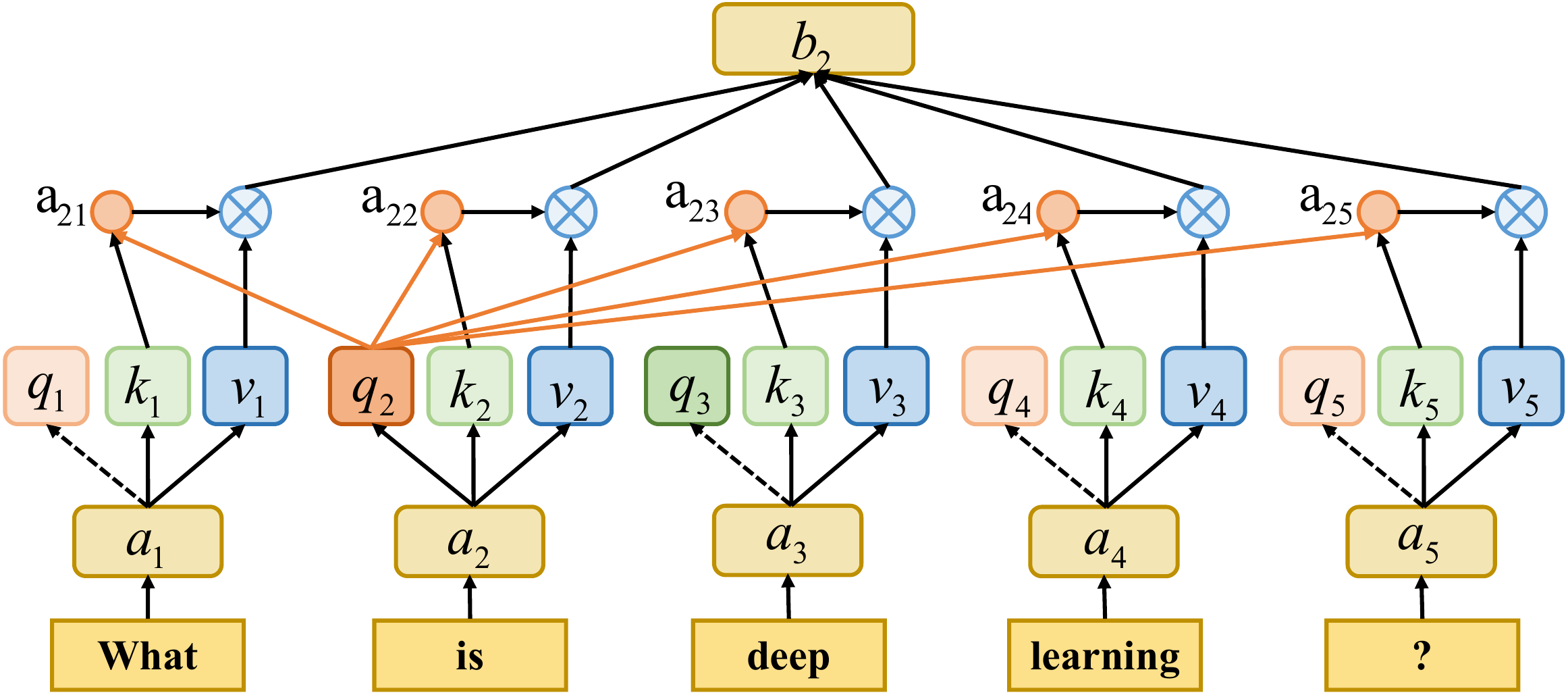}
    \caption{An example of self-attention for calculating output vector $b_{2}$.}
    \label{figureYang_Liu}
\end{figure}

Self-attention \cite{DBLP:conf/nips/VaswaniSPUJGKP17} captures the weight distribution of words in sentences by constructing K, Q and V matrices among sentences that can capture long-range dependencies on text classification. 
We give an example for self-attention, as shown in Fig.~\ref{figureYang_Liu}. 
Each input word vector $a_{i}$ can be represented as three n-dimensional vectors, including $q_{i}$, $k_{i}$ and $v_{i}$. 
After self-attention, the output vector $b_{i}$ can be represented as $ \sum_{j}softmax(a_{ij})v_{j}$ and $ a_{ij}=q_{i} \cdot k_{j} / \sqrt{n}$. All output vectors can be parallelly computed. 
Lin et al. \cite{DBLP:conf/iclr/LinFSYXZB17} used source token self-attention to explore the weight of every token to the entire sentence in the sentence representation task. 
To capture long-range dependencies, Bi-directional Block Self-Attention Network (Bi-BloSAN) \cite{DBLP:conf/iclr/ShenZL0Z18} uses an intra-block Self-Attention Network (SAN) to every block split by sequence and an inter-block SAN to the outputs.

Aspect-Based Sentiment Analysis (ABSA) \cite{inproceedings2017, DBLP:conf/aaai/MaPC18} breaks down a text into multiple aspects and allocates each aspect a sentiment polarity. 
The sentiment polarity can be divided into three types: positive, neutral and negative. Some attention-based models are proposed to identify the fine-grained opinion polarity towards a specific aspect for aspect-based sentiment tasks. 
ATAE-LSTM \cite{DBLP:conf/emnlp/WangHZZ16} can concentrate on different parts of each sentence according to the input through the attention mechanisms. 
MGAN \cite{DBLP:conf/emnlp/FanFZ18} presents a fine-grained attention mechanism with a coarse-grained attention mechanism to learn the word-level interaction between context and aspect.

To catch the complicated semantic relationship among each question and candidate answers for the QA task, Tan et al. \cite{tan-etal-2016-improved} introduce CNN and RNN and generate answer embeddings by using a simple one-way attention mechanism affected through the question context. 
The attention captures the dependence among the embeddings of questions and answers. 
Extractive QA can be seen as the text classification task. 
It inputs a question and multiple candidates answers and classifies every candidate answer to recognize the correct answer. 
Furthermore, AP-BILSTM \cite{DBLP:journals/corr/SantosTXZ16} with a two-way attention mechanism can learn the weights between the question and each candidate answer to obtain the importance of each candidate answer to the question.

\begin{figure*}[!htbp]
    \centering
    \includegraphics[width=\linewidth]{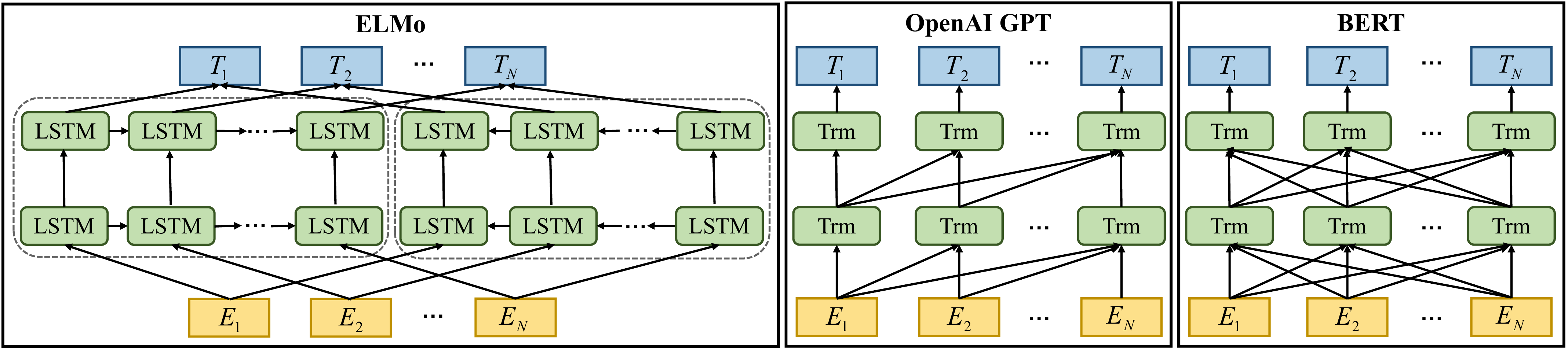}
    \caption{Differences in pre-trained model architectures \cite{DBLP:conf/naacl/DevlinCLT19}, including BERT, OpenAI GPT and ELMo. $E_{i}$ represents embedding of $i$ th input. Trm represents the transformer block. $T_{i}$ represents predicted tag of $i$ th input.}
    \label{figureComp}
\end{figure*}

\subsubsection{Pre-trained Methods}
 
Pre-trained language models \cite{DBLP:journals/corr/abs-2003-08271} effectively learn global semantic representation and significantly boost NLP tasks, including text classification. 
It generally uses unsupervised methods to mine semantic knowledge automatically and then construct pre-training targets so that machines can learn to understand semantics.

As shown in Fig.~\ref{figureComp}, we give differences in the model architectures among the Embedding from Language Model (ELMo) \cite{DBLP:conf/naacl/PetersNIGCLZ18}, OpenAI GPT \cite{Radford2018ImprovingLU}, and BERT \cite{DBLP:conf/naacl/DevlinCLT19}. 
ELMo \cite{DBLP:conf/naacl/PetersNIGCLZ18} is a deep contextualized word representation model, which is readily integrated into models. 
It can model complicated characteristics of words and learn different representations for various linguistic contexts. 
It learns each word embedding according to the context words with the bi-directional LSTM. 
GPT \cite{Radford2018ImprovingLU} employs supervised fine-tuning and unsupervised pre-training to learn general representations that transfer with limited adaptation to many NLP tasks. 
Furthermore, the domain of the target dataset does not need to be similar to the domain of unlabeled datasets. 
The training procedure of the GPT algorithm usually includes two stages. 
Firstly, the initial parameters of a neural network model are learned by a modeling objective on the unlabeled dataset. 
We can then employ the corresponding supervised objective to accommodate these parameters for the target task.  
To pre-train deep bidirectional representations from the unlabeled text through joint conditioning on both left and right context in every layer, BERT model \cite{DBLP:conf/naacl/DevlinCLT19}, proposed by Google, significantly improves performance on NLP tasks, including text classification. BERT applies the bi-directional encoder designed to pre-train the bi-directional representation of depth by jointly adjusting the context in all layers. It can utilize contextual information when predicting which words are masked.
It is fine-tuned by adding just an additional output layer to construct models for multiple NLP tasks, such as SA, QA, and machine translation.
Comparing with these three models, ELMo is a feature-based method using LSTM, and BERT and OpenAI GPT are fine-tuning approaches using Transformer. 
Furthermore, ELMo and BERT are bidirectional training models and OpenAI GPT is training from left to right. 
Therefore, BERT gets a better result, which combines the advantages of ELMo and OpenAI GPT.

Transformer-based models can parallelize computation without considering the sequential information suitable for large scale datasets, making it popular for NLP tasks.
Thus, some other works are used for text classification tasks and get excellent performance. 
RoBERTa \cite{DBLP:journals/corr/abs-1907-11692}, is an improved version of BERT, adopts the dynamic masking method that generates the masking pattern every time with a sequence to be fed into the model. 
It uses more data for longer pre-training and estimates the influence of various essential hyperparameters and the size of training data. To be specific: 1) The training time is longer (a total of nearly 200,000 training, nearly 1.6 billion training data have been seen), the batch size (8K) is larger, and the training data is more (30G Chinese training, including 300 million sentences and 10 billion words); 2) It removes the next sentence prediction (NSP) task; 3) It employs more extended training sequence; 4) It dynamically adjusts the masking mechanism and use the full word mask.

XLNet \cite{DBLP:conf/nips/YangDYCSL19} is a generalized autoregressive pre-training approach. 
Unlike BERT, the denoising autoencoder with the mask is not used in the first stage, but the autoregressive LM is used. It maximizes the expected likelihood across the whole factorization order permutations to learn the bidirectional context. 
Furthermore, it can overcome the weaknesses of BERT by an autoregressive formulation and integrate ideas from Transformer-XL \cite{DBLP:conf/acl/DaiYYCLS19} into pre-training.

BERT model has many parameters. In order to reduce the parameters, ALBERT \cite{DBLP:conf/iclr/LanCGGSS20} uses two-parameter simplification schemes. It reduces the fragmentation vector's length and shares parameters with all encoders. It also replaces the next sentence matching task with the next sentence order task and continuously blocks fragmentation. When the ALBERT model is pre-trained on a massive Chinese corpus, the parameters are less and better performance.
In general, these methods adopt unsupervised objective functions for pre-training, including the next sentence prediction, masking technology, and permutation. 
These target functions based on the word prediction demonstrate a strong ability to learn the word dependence and semantic structure \cite{DBLP:conf/acl/JawaharSS19}. 

\begin{figure}[!htbp]
    \centering
    \includegraphics[width=\linewidth]{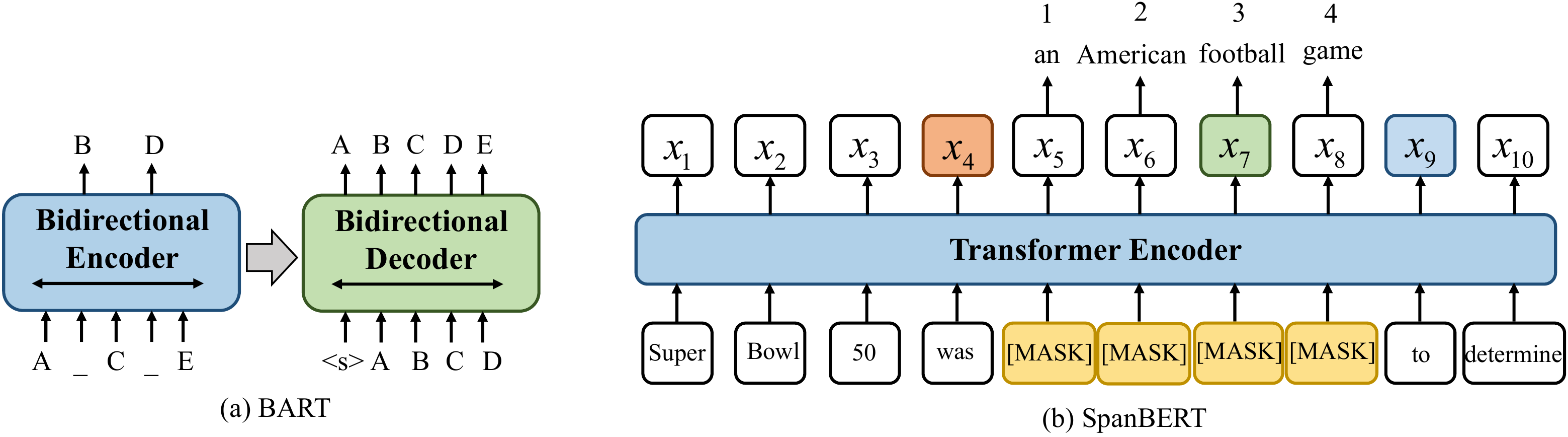}
    \caption{The architecture of BART \cite{DBLP:conf/acl/LewisLGGMLSZ20} and SpanBERT \cite{DBLP:journals/tacl/JoshiCLWZL20}.}
    \label{BART_SPANBERT}
\end{figure}  

BART \cite{DBLP:conf/acl/LewisLGGMLSZ20} is a denoising autoencoder based on the Seq2Seq model, as shown in Fig. \ref{BART_SPANBERT} (a). The pre-training of BART consists of two steps. Firstly, it uses a noise function to destroy the text. Secondly, the Seq2Seq model is used to reconstruct the original text. In various noise methods, by randomly shuffling the order of the original sentence and then using the first new text filling method to obtain optimal performance. The new text filling method is replacing the text fragment with a single mask token. It uses only a specific masked token to indicate that a token is masked.

SpanBERT \cite{DBLP:journals/tacl/JoshiCLWZL20} is specially designed to better represent and predict spans of text, as shown in Fig. \ref{BART_SPANBERT} (b). It optimizes BERT from three aspects and achieves good results in multiple tasks such as QA. The specific optimization is embodied in three aspects. Firstly, the span mask scheme is proposed to mask a continuous paragraph of text randomly. Secondly, Span Boundary Objective (SBO) is added to predict span by the token next to the span boundary to get the better performance to finetune stage. Thirdly, the NSP pre-training task is removed.

ERNIE \cite{DBLP:journals/corr/abs-1904-09223} is based on the method of knowledge enhancement. It learns the semantic relations in the real world by modeling the prior semantic knowledge such as entity concepts in massive datasets. Specifically, ERNIE enables the model to learn the semantic representation of complete concepts by masking semantic units such as words and entities. It mainly consists of a Transformer encoder and task embedding. In the Transformer encoder, the context information of each token is captured by the self-attention mechanism, and the context representation is generated for embedding. Task embedding is used for tasks with different characteristics.

\subsubsection{GNN-based Methods}

\begin{figure}[!htbp]
    \centering
    \includegraphics[width=0.95\linewidth]{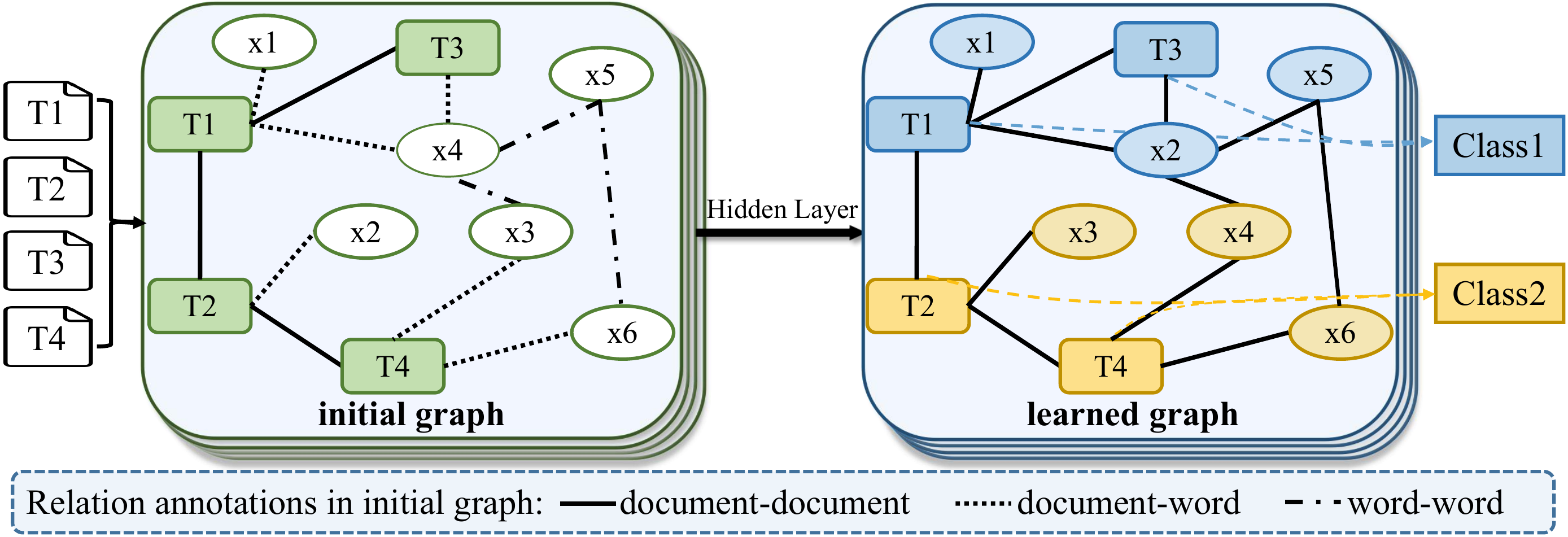}
    \caption{The GNN-based model. The initial graph differently depending on how the graph is designed. We give an example to establish edges between documents and documents, documents and sentences, and words to words.}
    \label{figure_GCN_SGC}
\end{figure}

The DNN models like CNN get great performance on regular structure, not for arbitrarily structured graphs. Some researchers study how to expand on arbitrarily structured graphs \cite{DBLP:conf/nips/DefferrardBV16, peng2021reinforced}. 
With the increasing attention of Graph Neural Networks (GNNs), GNN-based models \cite{peng2021lime, li2021higher} obtain excellent performance by encoding syntactic structure of sentences on semantic role labeling task \cite{DBLP:conf/emnlp/MarcheggianiT17}, relation classification task \cite{DBLP:journals/jamia/LiJL19} and machine translation task \cite{DBLP:conf/emnlp/BastingsTAMS17}. 
It turns text classification into a graph node classification task. We show a GCN model for text classification with four input texts, as shown in Fig.~\ref{figure_GCN_SGC}. Firstly, the four input texts $T=[T_{1}, T_{2}, T_{3}, T_{4}]$ and the words $X=[x_{1}, x_{2}, x_{3}, x_{4}, x_{5}, x_{6}]$ in the text, defined as nodes, are constructed into the graph structures. 
The graph nodes are connected by bold black edges, which indicates document-word edges and word-word edges. 
The weight of each word-word edge usually means their co-occurrence frequency in the corpus.
Then, the words and texts are represented through the hidden layer. Finally, the label of all input texts can be predicted by the graph. 

The GNN-based models can learn the syntactic structure of sentences, making some researchers study using GNN for text classification. 
DGCNN \cite{DBLP:conf/www/PengLHLBWS018} is a graph-CNN converting text to graph-of-words, having the advantage of learning different levels of semantics with CNN models. 
Yao et al. \cite{DBLP:conf/aaai/YaoM019} propose the Text Graph Convolutional Network (TextGCN), which builds a heterogeneous word text graph for a whole dataset and captures global word co-occurrence information. 
To enable GNN-based models to underpin online testing, Huang et al. \cite{DBLP:conf/emnlp/HuangMLZW19} build graphs for each text with global parameter sharing, not a corpus-level graph structure, to help preserve global information and reduce the burden. 
TextING \cite{DBLP:conf/acl/ZhangYCWWW20} builds individual graphs for each document and learns text-level word interactions by GNN to effectively produce embeddings for obscure words in the new text.

Graph ATtention network (GAT) \cite{DBLP:conf/iclr/VelickovicCCRLB18} employs masked self-attention layers by attending over its neighbors. 
Thus, some GAT-based models are proposed to compute the hidden representations of each node. 
The Heterogeneous Graph ATtention networks (HGAT) \cite{DBLP:conf/emnlp/HuYSJL19} with a dual-level attention mechanism learns the importance of different neighboring nodes and node types in the current node. 
The model propagates information on the graph and captures the relations to address the semantic sparsity for semi-supervised short text classification. 
MAGNET \cite{DBLP:conf/icaart/PalSS20} is proposed to capture the correlation among the labels based on GATs, which learns the crucial dependencies between the labels and generates classifiers by a feature matrix and a correlation matrix. 

Event Prediction (EP) can be divided into generated event prediction and selective event prediction (also known as script event prediction). 
EP, referring to scripted event prediction in this review, infers the subsequent event according to the existing event context. Unlike other text classification tasks, texts in EP are composed of a series of sequential subevents. 
Extracting features of the relationship among such subevents is of critical importance. 
SGNN \cite{DBLP:conf/ijcai/LiDL18} is proposed to model event interactions and learn better event representations by constructing an event graph to utilize the event network information better. 
The model makes full use of dense event connections for the EP task.

\subsubsection{Others}

In addition to all the above models, there are some other individual models. Here we introduce some exciting models.

\paragraph{Siamese Neural Network.}
The siamese neural network \cite{DBLP:conf/nips/BromleyGLSS93} is also called a twin neural network (Twin NN). 
It utilizes equal weights while working in tandem using two distinct input vectors to calculate comparable output vectors. 
Mueller et al. \cite{DBLP:conf/aaai/MuellerT16} present a siamese adaptation of the LSTM network comprised of couples of variable-length sequences. 
The model is employed to estimate the semantic similarity among texts, exceeding carefully handcrafted features and proposed neural network models of higher complexity. 
The model further represents text employing neural networks whose inputs are word vectors learned separately from a vast dataset.
To settle unbalanced data classification in the medical domain, Jayadeva et al. \cite{JAYADEVA201934} use a Twin NN model to learn from enormous unbalanced corpora. 
The objective functions achieve the Twin SVM approach with non-parallel decision boundaries for the corresponding classes, and decrease the Twin NN complexity, optimizing the feature map to better discriminate among classes.

\paragraph{Virtual Adversarial Training (VAT)}
Deep learning methods require many extra hyperparameters, which increase the computational complexity.
VAT \cite{miyato2015distributional}, regularization based on local distributional smoothness can be used in semi-supervised tasks, requires only some hyperparameters, and can be interpreted directly as robust optimization. 
Miyato et al. \cite{DBLP:conf/iclr/MiyatoDG17} use VAT to effectively improve the robustness and generalization ability of the model and word embedding performance.

\paragraph{Reinforcement Learning (RL)}
RL learns the best action in a given environment through maximizing cumulative rewards. 
Zhang et al. \cite{DBLP:conf/aaai/ZhangHZ18} offer an RL approach to establish structured sentence representations via learning the structures related to tasks. 
The model has Information Distilled LSTM (ID-LSTM) and Hierarchical Structured LSTM (HS-LSTM) representation models.
The ID-LSTM learns the sentence representation by choosing essential words relevant to tasks, and the HS-LSTM is a two-level LSTM for modeling sentence representation.

\paragraph{Memory Networks}
Memory networks \cite{weston2015memory} learn to combine the inference components and the long-term memory component. 
Li et al. \cite{DBLP:conf/emnlp/LiL17} use two LSTMs with extended memories and neural memory operations for jointly handling the extraction tasks of aspects and opinions via memory interactions. 
Topic Memory Networks (TMN) \cite{DBLP:conf/emnlp/ZengLSGLK18} is an end-to-end model that encodes latent topic representations indicative of class labels.

\paragraph{QA Style for Sentiment Classification Task.}
It is an interesting attempt to treat the sentiment classification task as a QA task. 
Shen et al. \cite{DBLP:conf/emnlp/ShenSWKLLSZZ18} create a high-quality annotated corpus. 
A three-stage hierarchical matching network was proposed to consider the matching information between questions and answers.

\paragraph{External Commonsense Knowledge.}
Due to the insufficient information of the event itself to distinguish the event for the EP task, Ding et al. \cite{DBLP:conf/emnlp/DingLLLD19} consider that the event extracted from the original text lacked common knowledge, such as the intention and emotion of the event participants. The model improves the effect of stock prediction, EP, and so on.

\paragraph{Quantum Language Model.}
In the quantum language model, the words and dependencies among words are represented through fundamental quantum events. 
Zhang et al. \cite{Zhang2019A} design a quantum-inspired sentiment representation method to learn both the semantic and the sentiment information of subjective text. 
By inputting density matrices to the embedding layer, the performance of the model improves. 

\textbf{Summary.}
RNN computes sequentially and cannot be calculated in parallel. The shortcoming of RNN makes it more challenging to become mainstream in the current trend that models tend to have deeper and more parameters. CNN extracts features from text vectors through the convolution kernel. The number of features captured by the convolution kernel is related to its size. CNN is deep enough that, in theory, it can capture features at long distances. Due to insufficient optimization methods for parameters of the deep network and the loss of location information due to the pooling layer, the deeper layer does not bring significant improvement. Compared with RNN, CNN has parallel computing capability and can effectively retain location information for the improved version of CNN. Still, it has weak feature capture capability for long-distance. GNN builds a graph for text. When a valid graph structure is designed, the learned representation can better capture the structural information. Transformer treats the input text as a fully connected graph, with attention score weights on the edges. It is capable of parallel computing and is highly efficient in extracting features between different words by self-attention, solving short-term memory problems. However, the attention mechanism in Transformer is computation-heavy, especially when dealing with long sequences. Some improved models \cite{DBLP:conf/iclr/LanCGGSS20, DBLP:conf/nips/ZafrirBIW19} for computing complexity in Transformer have recently been proposed. Overall, Transformer is a better choice for text classification.
Deep Learning consists of multiple hidden layers in a neural network with a higher level of complexity and can be trained on unstructured data. 
Deep learning can learn language features and master higher level and more abstract language features based on words and vectors. 
Deep learning architecture can learn feature representations directly from the input without too many manual interventions and prior knowledge.
However, deep learning technology is a data-driven method that requires enormous data to achieve high performance. 
Although self-attention based models can bring some interpretability among words for DNNs, it is not enough comparing with traditional models to explain why and how it works well.

\section{Datasets and Evaluation Metrics} \label{Section 4}

\begin{table*}[!htbp]
\centering
\caption{Summary statistics for the datasets. C: Number of target classes. L: Average sentence length. N: Dataset size.}
\label{tab:datasets}
\resizebox{\textwidth}{!}{
    \begin{tabular}{c|rrrclcc}
    \toprule
\textbf{Datasets} & \textbf{\#C} &\textbf{\#L} & \textbf{\#N} & \textbf{Language} & \textbf{Related Papers} & \textbf{Sources} &\textbf{Applications}\\ 
\midrule
MR & 2 & 20 & 10,662 & English& \cite{DBLP:conf/emnlp/Kim14, DBLP:conf/acl/KalchbrennerGB14, DBLP:conf/emnlp/YangZYLZZ18, DBLP:conf/aaai/YaoM019} & \cite{movie-review-data} &SA \\ \hline

SST-1 & 5 & 18 & 11,855 & English& \cite{DBLP:conf/emnlp/SocherPWCMNP13, DBLP:conf/emnlp/Kim14}  \cite{DBLP:conf/acl/TaiSM15, DBLP:conf/icml/ZhuSG15}\cite{DBLP:conf/emnlp/0001DL16} &\cite{sentiment} &SA \\ \hline

SST-2 & 2 & 19 & 9,613& English & \cite{DBLP:conf/emnlp/SocherPWCMNP13, DBLP:conf/emnlp/Kim14, DBLP:conf/emnlp/LiuQCWH15}  \cite{DBLP:conf/ijcai/LiuQH16, DBLP:conf/naacl/DevlinCLT19} & \cite{socher-etal-2013-recursive} &SA\\ \hline

MPQA & 2 & 3 & 10,606 & English& \cite{DBLP:conf/emnlp/SocherPHNM11, DBLP:conf/emnlp/Kim14, DBLP:conf/iclr/ShenZL0Z18} & \cite{mpqa} &SA\\ \hline

IMDB & 2 & 294 & 50,000 & English& 
\cite{DBLP:conf/acl/IyyerMBD15}\cite{DBLP:conf/naacl/YangYDHSH16}
\cite{DBLP:conf/emnlp/LiuQCWH15}  \cite{DBLP:conf/ijcai/LiuQH16}  \cite{DBLP:conf/iclr/MiyatoDG17}  & \cite{DBLP:conf/kdd/DiaoQWSJW14} &SA\\ \hline

Yelp.P & 2 & 153 & 598,000 & English& \cite{DBLP:conf/nips/ZhangZL15, DBLP:conf/acl/JohnsonZ17} & \cite{DBLP:conf/emnlp/TangQL15} &SA\\ \hline

Yelp.F & 5 & 155 & 700,000& English & \cite{DBLP:conf/nips/ZhangZL15, DBLP:conf/naacl/YangYDHSH16, DBLP:conf/acl/JohnsonZ17} & \cite{DBLP:conf/emnlp/TangQL15} &SA\\ \hline
Amz.P & 2 & 91 & 4,000,000 & English& \cite{DBLP:conf/nips/YouZWDMZ19, DBLP:conf/nips/ZhangZL15} & \cite{amazon-review} &SA\\ \hline

Amz.F & 5 & 93 & 3,650,000& English & \cite{DBLP:conf/nips/ZhangZL15, DBLP:conf/naacl/YangYDHSH16, DBLP:conf/nips/YouZWDMZ19} & \cite{amazon-review} &SA\\ \hline

Twitter & 3 & 19 & 11,209& English & \cite{DBLP:conf/acl/KalchbrennerGB14}\cite{DBLP:conf/ijcai/WangWZY17} &\cite{task2} &SA\\ \hline

NLP\&CC 2013&2&-&115,606& Multi-language&\cite{DBLP:conf/emnlp/ZhouWX16}& \cite{tcci.ccf.org.cn} &SA\\ \hline

20NG & 20 & 221 & 18,846 & English& \cite{DBLP:conf/aaai/LaiXLZ15, DBLP:conf/icml/JohnsonZ16, DBLP:conf/iclr/BaoWCB20, DBLP:conf/aaai/YaoM019, DBLP:conf/icml/WuSZFYW19}  & \cite{datasets-for-single-label-textcategorization} &NC \\ \hline

AG News & 4 & 45/7 &127,600 & English&  \cite{DBLP:conf/acl/JohnsonZ17, DBLP:conf/ijcai/WangWZY17}
\cite{DBLP:conf/emnlp/YangZYLZZ18, DBLP:conf/nips/YangDYCSL19}    & \cite{AG-News} &NC\\ \hline

R8 & 8 & 66&7,674 & English& \cite{DBLP:conf/aaai/YaoM019, DBLP:conf/icml/WuSZFYW19}  \cite{DBLP:conf/emnlp/HuangMLZW19}  & \cite{textmining} &NC\\ \hline

R52 & 52 & 70 & 9,100 & English& \cite{DBLP:conf/aaai/YaoM019, DBLP:conf/icml/WuSZFYW19}  \cite{DBLP:conf/emnlp/HuangMLZW19} & \cite{textmining} &NC\\ \hline

Sogou & 6 & 578 & 510,000& Chinese & \cite{DBLP:conf/nips/ZhangZL15} &\cite{DBLP:conf/www/WangZMR08} &NC \\ \hline

Newsgroup&20 & - &  18,846& English&\cite{DBLP:conf/cikm/LiLCOL18}&\cite{DBLP:conf/cikm/LiLCOL18}&NC\\ \hline

DBPedia& 14 &55 & 630,000 & English& \cite{DBLP:conf/nips/ZhangZL15, DBLP:conf/acl/JohnsonZ17, DBLP:conf/iclr/MiyatoDG17, DBLP:conf/cncl/SunQXH19} &\cite{DBLP:journals/semweb/LehmannIJJKMHMK15} &TL \\ \hline

Ohsumed & 23 &136 &7,400& English &\cite{DBLP:conf/aaai/YaoM019, DBLP:conf/icml/WuSZFYW19, DBLP:conf/emnlp/HuangMLZW19}  & \cite{ohsumed} &TL\\ \hline

YahooA & 10 & 112 & 1,460,000& English & \cite{DBLP:conf/nips/ZhangZL15, DBLP:conf/naacl/YangYDHSH16}&  \cite{DBLP:conf/nips/ZhangZL15} &TL\\ \hline

EUR-Lex&3,956&1,239&19,314& English& \cite{DBLP:conf/sigir/LiuCWY17}  \cite{DBLP:conf/acl/ChalkidisFMA19, peng2019hierarchical}  \cite{DBLP:conf/acl/ChalkidisFMA19}&\cite{eurlex} &TL\\ \hline

Amazon670K & 670 &244& 643,474& English&\cite{DBLP:conf/emnlp/ShimuraLF18, DBLP:conf/nips/YouZWDMZ19}&\cite{XMLRepository} &TL\\ \hline
Google news &152 & 6 &  11,109& English&\cite{DBLP:conf/kdd/YinW14, DBLP:journals/apin/ChenGL20, 9152157}&\cite{DBLP:conf/kdd/YinW14}&TL\\ \hline

TweetSet 2011-2012 &89 &- &  2,472& English&\cite{DBLP:conf/kdd/YinW14, 9152157}&\cite{DBLP:conf/kdd/YinW14}&TL\\ \hline

TweetSet 2011-2015&269 & 8 & 30,322 & English&\cite{DBLP:journals/isci/ChenGL19, DBLP:journals/apin/ChenGL20}&\cite{DBLP:journals/isci/ChenGL19}&TL\\ \hline
Bing &4  & 20 & 34,871& English&\cite{DBLP:conf/ijcai/WangWZY17} &\cite{DBLP:conf/cikm/WangWLW14}&TL\\ \hline

Fudan&20&2981&18,655& Chinese&\cite{DBLP:conf/aaai/LaiXLZ15}&\cite{datatang}&TL\\ \hline

SQuAD & - & 5,000 &  5,570& English &\cite{DBLP:conf/naacl/PetersNIGCLZ18, DBLP:conf/naacl/PetersNIGCLZ18, DBLP:journals/corr/abs-1907-11692, DBLP:conf/iclr/LanCGGSS20}  & \cite{DBLP:conf/emnlp/RajpurkarZLL16}&QA \\ \hline

TREC-QA & - & 1,162 & 68& English & \cite{DBLP:journals/corr/SantosTXZ16}&\cite{DBLP:conf/naacl/YaoDCC13} &QA \\ \hline

TREC&6&10&5,952& English&\cite{DBLP:conf/emnlp/Kim14, DBLP:conf/acl/KalchbrennerGB14, DBLP:conf/emnlp/LiuQCWH15}  \cite{DBLP:conf/ijcai/WangWZY17}&\cite{QC} &QA \\\hline

WikiQA & - &  873 &  243& English &\cite{DBLP:conf/emnlp/YangYM15, DBLP:journals/corr/SantosTXZ16}  & \cite{DBLP:conf/emnlp/YangYM15}&QA \\ \hline

Subj & 2 & 23 & 10,000& English &\cite{DBLP:conf/emnlp/Kim14, DBLP:conf/ijcai/LiuQH16, DBLP:conf/emnlp/YangZYLZZ18} & \cite{pang-lee-2004-sentimental}&QA \\ \hline

CR & 2 & 19 & 3,775 & English& \cite{DBLP:conf/emnlp/Kim14, DBLP:conf/emnlp/YangZYLZZ18} & \cite{DBLP:conf/kdd/HuL04}&QA \\ \hline

Reuters & 90 & 168 & 10,788 & English& \cite{DBLP:conf/emnlp/YangZYLZZ18, DBLP:conf/icaart/PalSS20}& \cite{nlp-reuters}&ML \\ \hline

Reuters10 & 10 & 168 & 9,979& English & \cite{DBLP:journals/ijon/KimJPC20}& \cite{nlp-reuters10}&ML \\ \hline

RCV1 & 103 & 240 &807,595& English& \cite{DBLP:conf/icml/JohnsonZ16, DBLP:conf/emnlp/ShimuraLF18, DBLP:conf/www/PengLHLBWS018, DBLP:conf/acl/ChalkidisFMA19}  & \cite{DBLP:journals/jmlr/LewisYRL04} &ML\\ \hline

RCV1-V2 & 103 & 124 & 804,414 & English& \cite{DBLP:conf/coling/YangSLMWW18, DBLP:conf/icaart/PalSS20} & \cite{lyrl2004_rcv1v2_README.htm} &ML\\ \hline

AAPD & 54 & 163 & 55,840 & English&\cite{DBLP:conf/coling/YangSLMWW18, DBLP:conf/icaart/PalSS20}  & \cite{SGM}&ML \\ 
\bottomrule
    \end{tabular}}
\end{table*}

\subsection{Datasets}

The availability of labeled datasets for text classification has become the main driving force behind the fast advancement of this research field.
In this section, we summarize the characteristics of these datasets in terms of domains and give an overview in Table ~\ref{tab:datasets}, including the number of categories, average sentence length, the size of each dataset, related papers, data sources to access and applications.
 
\subsubsection{Sentiment Analysis (SA)}
SA is the process of analyzing and reasoning the subjective text within emotional color. 
It is crucial to get information on whether it supports a particular point of view from the text that is distinct from the traditional text classification that analyzes the objective content of the text. 
SA can be binary or multi-class. 
Binary SA is to divide the text into two categories, including positive and negative. 
Multi-class SA classifies text to multi-level or fine-grained labels. 
The SA datasets include Movie Review (MR) \cite{DBLP:conf/acl/PangL05, movie-review-data}, Stanford Sentiment Treebank (SST) \cite{sentiment}, Multi-Perspective Question Answering (MPQA) \cite{DBLP:journals/lre/WiebeWC05, mpqa}, IMDB \cite{DBLP:conf/kdd/DiaoQWSJW14}, Yelp \cite{DBLP:conf/emnlp/TangQL15}, Amazon Reviews (AM) \cite{DBLP:conf/nips/ZhangZL15}, NLP\&CC 2013 \cite{tcci.ccf.org.cn}, Subj \cite{pang-lee-2004-sentimental}, CR \cite{DBLP:conf/kdd/HuL04}, SS-Twitter \cite{DBLP:journals/jasis/ThelwallBP12}, SS-Youtube \cite{DBLP:journals/jasis/ThelwallBP12}, SE1604 \cite{Nakov2016SemEval} and so on.
Here we detail several datasets.

\textbf{MR.} The MR is a movie review dataset, each of which corresponds to a sentence. The corpus has 5,331 positive data and 5,331 negative data. 10-fold cross-validation by random splitting is commonly used to test MR.

\textbf{SST.} The SST is an extension of MR. 
It has two categories. SST-1 with fine-grained labels with five classes. 
It has 8,544 training texts and 2,210 test texts, respectively. Furthermore, SST-2 has 9,613 texts with binary labels being partitioned into 6,920 training texts, 872 development texts, and 1,821 testing texts.

\textbf{MPQA.} The MPQA is an opinion dataset. It has two class labels and also an MPQA dataset of opinion polarity detection sub-tasks. MPQA includes 10,606 sentences extracted from news articles from various news sources. It should be noted that it contains 3,311 positive texts and 7,293 negative texts without labels of each text.

\textbf{IMDB reviews.} The IMDB review is developed for binary sentiment classification of film reviews with the same amount in each class. It can be separated into training and test groups on average, by 25,000 comments per group.

\textbf{Yelp reviews.} The Yelp review is summarized from the Yelp Dataset Challenges in 2013, 2014, and 2015. This dataset has two categories. Yelp-2 of these were used for negative and positive emotion classification tasks, including 560,000 training texts and 38,000 test texts. Yelp-5 is used to detect fine-grained affective labels with 650,000 training and 50,000 test texts in all classes.

\textbf{AM.} The AM is a popular corpus formed by collecting Amazon website product reviews \cite{amazon-review}. This dataset has two categories. The Amazon-2 with two classes includes 3,600,000 training sets and 400,000 testing sets. Amazon-5, with five classes, includes 3,000,000 and 650,000 comments for training and testing.

\subsubsection{News Classification (NC)}
News content is one of the most crucial information sources which has a critical influence on people. 
The NC system facilitates users to get vital knowledge in real-time. 
News classification applications mainly encompass: recognizing news topics and recommending related news according to user interest. 
The news classification datasets include 20 Newsgroups (20NG) \cite{datasets-for-single-label-textcategorization}, AG News (AG) \cite{DBLP:conf/nips/ZhangZL15, AG-News}, R8 \cite{textmining}, R52 \cite{textmining}, Sogou News (Sogou) \cite{DBLP:conf/cncl/SunQXH19} and so on. Here we detail several datasets.

\textbf{20NG.} The 20NG is a newsgroup text dataset. It has 20 categories with the same number of each category and includes 18,846 texts.

\textbf{AG.} The AG News is a search engine for news from academia, choosing the four largest classes. It uses the title and description fields of each news. AG contains 120,000 texts for training and 7,600 texts for testing.

\textbf{R8 and R52.} R8 and R52 are two subsets which are the subset of Reuters \cite{nlp-reuters}. R8 has 8 categories, divided into 2,189 test files and 5,485 training courses. R52 has 52 categories, split into 6,532 training files and 2,568 test files. 

\textbf{Sogou.} The Sogou combines two datasets, including SogouCA and SogouCS news sets. The label of each text is the domain names in the URL.

\subsubsection{Topic Labeling (TL)}
The topic analysis attempts to get the meaning of the text by defining the sophisticated text theme. 
The topic labeling is one of the essential components of the topic analysis technique, intending to assign one or more subjects for each document to simplify the topic analysis. 
The topic labeling datasets include DBPedia \cite{DBLP:journals/semweb/LehmannIJJKMHMK15}, Ohsumed \cite{ohsumed}, Yahoo answers (YahooA) \cite{DBLP:conf/nips/ZhangZL15}, EUR-Lex \cite{eurlex}, Amazon670K \cite{XMLRepository}, Bing \cite{DBLP:conf/cikm/WangWLW14}, Fudan \cite{datatang}, and PubMed \cite{DBLP:journals/biodb/Lu11}. Here we detail several datasets.

\textbf{DBpedia.} The DBpedia is a large-scale multi-lingual knowledge base generated using Wikipedia's most ordinarily used infoboxes. It publishes DBpedia each month, adding or deleting classes and properties in every version. DBpedia's most prevalent version has 14 classes and is divided into 560,000 training data and 70,000 test data.

\textbf{Ohsumed.} The Ohsumed belongs to the MEDLINE database. It includes 7,400 texts and has 23 cardiovascular disease categories. All texts are medical abstracts and are labeled into one or more classes.

\textbf{YahooA.} The YahooA is a topic labeling task with 10 classes. It includes 140,000 training data and 5,000 test data. All text contains three elements, being question titles, question contexts, and best answers, respectively.

\subsubsection{Question Answering (QA)}
The QA task can be divided into two types: the extractive QA and the generative QA. 
The extractive QA gives multiple candidate answers for each question to choose which one is the right answer. 
Thus, the text classification models can be used for the extractive QA task.
The QA discussed in this paper is all extractive QA. 
The QA system can apply the text classification model to recognize the correct answer and set others as candidates. 
The question answering datasets include Stanford Question Answering Dataset (SQuAD) \cite{DBLP:conf/emnlp/RajpurkarZLL16}, TREC-QA \cite{QC}, WikiQA \cite{DBLP:conf/emnlp/YangYM15}, Subj \cite{pang-lee-2004-sentimental}, CR \cite{DBLP:conf/kdd/HuL04}, MS MARCO \cite{DBLP:conf/nips/NguyenRSGTMD16}, and Quora \cite{QuestionPairs}. Here we detail several datasets.

\textbf{SQuAD.} The SQuAD is a set of question and answer pairs obtained from Wikipedia articles. The SQuAD has two categories. SQuAD1.1 contains 536 pairs of 107,785 Q\&A items. SQuAD2.0 combines 100,000 questions in SQuAD1.1 with more than 50,000 unanswerable questions that crowd workers face in a form similar to answerable questions \cite{DBLP:conf/acl/RajpurkarJL18}.

\textbf{TREC-QA.} The TREC-QA includes 5,452 training texts and 500 testing texts. It has two versions. TREC-6 contains 6 categories, and TREC-50 has 50 categories.

\textbf{WikiQA.} The WikiQA dataset includes questions with no correct answer, which needs to evaluate the answer. 

\textbf{MS MARCO.} The MS MARCO contains questions and answers. The questions and part of the answers are sampled from actual web texts by the Bing search engine. Others are generative. It is used for developing generative QA systems released by Microsoft.

\subsubsection{Natural Language Inference (NLI)}
NLI is used to predict whether the meaning of one text can be deduced from another. Paraphrasing is a generalized form of NLI. It uses the task of measuring the semantic similarity of sentence pairs to decide whether one sentence is the interpretation of another. The NLI datasets include Stanford Natural Language Inference (SNLI) \cite{DBLP:conf/emnlp/BowmanAPM15}, Multi-Genre Natural Language Inference (MNLI) \cite{DBLP:conf/naacl/WilliamsNB18}, Sentences Involving Compositional Knowledge (SICK) \cite{DBLP:conf/semeval/MarelliBBBMZ14}, Microsoft Research Paraphrase (MSRP) \cite{DBLP:conf/coling/DolanQB04}, Semantic Textual Similarity (STS) \cite{DBLP:journals/corr/abs-1708-00055}, Recognising Textual Entailment (RTE) \cite{DBLP:conf/mlcw/DaganGM05}, SciTail \cite{DBLP:conf/aaai/KhotSC18}, etc. Here we detail several of the primary datasets.

\textbf{SNLI.} The SNLI is generally applied to NLI tasks. It contains 570,152 human-annotated sentence pairs, including training, development, and test sets, which are annotated with three categories: neutral, entailment, and contradiction.

\textbf{MNLI.} The MNLI is an expansion of SNLI, embracing a broader scope of written and spoken text genres. It includes 433,000 sentence pairs annotated by textual entailment labels.

\textbf{SICK.} The SICK contains almost 10,000 English sentence pairs. It consists of neutral, entailment and contradictory labels.

\textbf{MSRP.} The MSRP consists of sentence pairs, usually for the text-similarity task. Each pair is annotated by a binary label to discriminate whether they are paraphrases. It respectively includes 1,725 training and 4,076 test sets.

\subsubsection{Multi-Label (ML) datasets}
In multi-label classification, an instance has multiple labels, and each label can only take one of the multiple classes. 
There are many datasets based on multi-label text classification. 
It includes Reuters \cite{nlp-reuters}, Reuters Corpus Volume I (RCV1) \cite{DBLP:journals/jmlr/LewisYRL04}, RCV1-2K \cite{DBLP:journals/jmlr/LewisYRL04}, Arxiv Academic Paper Dataset (AAPD) \cite{SGM}, Patent, Web of Science (WOS-11967) \cite{DBLP:conf/icmla/KowsariBHMGB17}, AmazonCat-13K \cite{b1FRNnCLFL}, BlurbGenreCollection (BGC) \cite{blurb-genre-collection}, etc. 
Here we detail several datasets.

\textbf{Reuters.} The Reuters is a popularly used dataset for text classification from Reuters financial news services. It has 90 training classes, 7,769 training texts, and 3,019 testing texts, containing multiple labels and single labels. There are also some Reuters sub-sets of data, such as R8, BR52, RCV1, and RCV1-v2.

\textbf{RCV1 and RCV1-2K.} The RCV1 is collected from Reuters News articles from 1996-1997, which is human-labeled with 103 categories. It consists of 23,149 training and 784,446 testing texts, respectively. The RCV1-2K dataset has the same features as the RCV1. However, the label set of RCV1-2K has been expanded with some new labels. It contains 2456 labels.

\textbf{AAPD.} The AAPD is a large dataset in the computer science field for the multi-label text classification from website \footnote{https://arxiv.org/}. It has 55,840 papers, including the abstract and the corresponding subjects with 54 labels in total. The aim is to predict the corresponding subjects of each paper according to the abstract.

\textbf{Patent Dataset.} The Patent Dataset is obtained from USPTO \footnote{https://www.uspto.gov/}, which is a patent system grating U.S. patents containing textual details such title and abstract. It contains 100,000 US patents awarded in the real-world with multiple hierarchical categories.

\textbf{WOS-11967.} The WOS-11967 is crawled from the Web of Science, consisting of abstracts of published papers with two labels for each example. It is shallower, but significantly broader, with fewer classes in total.

\subsubsection{Others}
There are some datasets for other applications, such as SemEval-2010 Task 8 \cite{DBLP:conf/naacl/HendrickxKKNSPP09}, ACE 2003-2004 \cite{DBLP:conf/lrec/StrasselPPSM08}, TACRED \cite{DBLP:conf/emnlp/ZhangZCAM17}, and NYT-10 \cite{DBLP:conf/pkdd/RiedelYM10}, FewRel \cite{FewRel}, Dialog State Tracking Challenge 4 (DSTC 4) \cite{DBLP:conf/iwsds/KimDBWH16}, ICSI Meeting Recorder Dialog Act (MRDA) \cite{DBLP:conf/icassp/AngLS05}, and Switchboard Dialog Act (SwDA) \cite{manualarticle}, and so on.

\subsection{Evaluation Metrics}

In terms of evaluating text classification models, accuracy and F1 score are the most used to assess the text classification methods. 
Later, with the increasing difficulty of classification tasks or the existence of some particular tasks, the evaluation metrics are improved. 
For example, evaluation metrics such as $P@K$ and $Micro\!-\!\!F1$ are used to evaluate multi-label text classification performance, and MRR is usually used to estimate the performance of QA tasks. 
In Table ~\ref{tab:Metrics}, we give the notations used in evaluation metrics.

\begin{table}
    \centering
    \caption{The notations used in evaluation metrics.}\label{tab:Metrics}
    \resizebox{240pt}{!}{
    \begin{tabular}{cc}
    \toprule
        \textbf{Notations} & \textbf{Descriptions} \\ 
        \midrule
        $TP$ & true positive \\ \hline
        $FP$ & false positive \\ \hline
        $TN$ & true negative \\ \hline
        $FN$ & false negative \\ \hline
        $TP_{t}$ & true positive of the $t$ th label on a text \\ \hline
        $FP_{t}$ & false positive of the $t$ th label on a text\\ \hline
        $TN_{t}$ & true negative of the $t$ th label on a text\\ \hline
        $FN_{t}$ & false negative of the $t$ th label on a text\\ \hline 
        $\mathcal{S}$& label set of all samples\\ \hline
        ${Q}$ & the number of predicted labels on each text \\ 
        \bottomrule
    \end{tabular}
    }
\end{table}

\subsubsection{Single-label metrics}Single-label text classification divides the text into one of the most likely categories applied in NLP tasks such as QA, SA, and dialogue systems \cite{DBLP:conf/naacl/LeeD16}. 
For single-label text classification, one text belongs to just one catalog, making it possible not to consider the relations among labels. 
Here, we introduce some evaluation metrics used for single-label text classification tasks.

\myparagraph{Accuracy and ErrorRate} The Accuracy and ErrorRate are the fundamental metrics for a text classification model. 
The $Accuracy$ and $ErrorRate$ are respectively defined as
\begin{equation}
Accuracy =\frac{(TP+TN)}{N},
\end{equation}
\begin{equation}
\quad ErrorRate = 1 - Accuracy =\frac{(FP+FN)}{N}. 
\end{equation}

\myparagraph{Precision, Recall and F1}
These are vital metrics utilized for unbalanced test sets, regardless of the standard type and error rate. 
For example, most of the test samples have a class label. 
$F1$ is the harmonic average of $Precision$ and $Recall$. 
$Precision$, $Recall$, and $F1$ as defined
\begin{equation}
Precision =\frac{TP}{TP+FP}, \quad
Recall =\frac{TP}{TP+FN},
\end{equation}
\begin{equation}
F1 =\frac{2 Precision \times Recall }{ Precision +Recall}.
\end{equation}
The desired results will be obtained when the accuracy, $F1$ and $Recall$ value reach 1. 
On the contrary, when the values become 0, the worst result is obtained. 
For the multi-class classification problem, the precision and recall value of each class can be calculated separately, and then the performance of the individual and whole can be analyzed.

\myparagraph{Exact Match (EM)} The EM \cite{A1977Maximum} is a metric for QA tasks, measuring the prediction that matches all the ground-truth answers precisely. It is the primary metric utilized on the SQuAD dataset.

\myparagraph{Mean Reciprocal Rank (MRR)} The MRR \cite{DBLP:conf/sigir/SeverynM15} is usually applied for assessing the performance of ranking algorithms on QA and Information Retrieval (IR) tasks. $MRR$ is defined as
\begin{equation}
MRR=\frac{1}{Q} \sum_{i=1}^{Q} \frac{1}{{rank}(i)},
\end{equation}
where ${rank}(i)$ is the ranking of the ground-truth answer at answer $i$-th.

\myparagraph{Hamming-Loss (HL)} The HL \cite{DBLP:journals/ml/SchapireS99} assesses the score of misclassified instance-label pairs where a related label is omitted or an unrelated is predicted.

Among these single-label evaluation metrics, the Accuracy is the earliest metric that calculates the proportion of the sample size that is predicted correctly and is not considered whether the predicted sample is a positive or a negative sample. $Precision$ calculates how many of the positive samples are actually positive, and the $Recall$ calculates how many of the positive examples in the sample are predicted correctly. Furthermore, $F1$ is the harmonic average of them, which is the most commonly used evaluation metrics. 

\subsubsection{Multi-label metrics}Compared with single-label text classification, multi-label text classification divides the text into multiple category labels, and the number of category labels is variable.
These metrics are designed for single label text classification, which are not suitable for multi-label tasks. 
Thus, there are some metrics designed for multi-label text classification.

\myparagraph{\bm{$ Micro\!-\!\!F1$}} The $ Micro\!-\!\!F1$ \cite{DBLP:books/daglib/0021593} is a measure that considers the overall accuracy and recall of all labels. The $ Micro\!-\!\!F1$ is defined as:
\begin{equation}
Micro\!-\!\!F1=\frac{2 {P}_{t} \times R_{t}}{{P}+{R}},
\end{equation}
where:
\begin{equation}
 P=\frac{\sum_{t \in \mathcal{S}} T P_{t}}{\sum_{t \in S} T P_{t}+F P_{t}},\quad  R=\frac{\sum_{t \in S} T P_{t}}{\sum_{t \in \mathcal{S}} T P_{t}+F N_{t}}.
\end{equation}

\myparagraph{\bm{$ Macro\!-\!\!F1$}} The $ Macro\!-\!\!F1$  \cite{DBLP:books/daglib/0021593} calculates the average $F1$ of all labels. Unlike $ Micro\!-\!\!F1$, which sets even weight to every example, $ Macro\!-\!\!F1$ sets the same weight to all labels in the average process.
Formally, $ Macro\!-\!\!F1$ is defined as:
\begin{equation}
{Macro}\!-\!\!F1=\frac{1}{\mathcal{S}} \sum_{t \in \mathcal{S}} \frac{2 {P}_{t} \times R_{t}}{{P_{t}}+{R_{t}}},
\end{equation}
where:
\begin{equation}
P_{t}=\frac{T P_{t}}{T P_{t}+F P_{t}},\quad R_{t}=\frac{T P_{t}}{T P_{t}+F N_{t}}.
\end{equation}
In addition to the above evaluation metrics, there are some rank-based evaluation metrics for extreme multi-label classification tasks, including $P@K$ and $NDCG@K$.

\myparagraph{Precision at
Top K (P@K)} The $P@K$   \cite{DBLP:conf/sigir/LiuCWY17} is the precision at the top k. For $P@K$, each text has a set of $\mathcal{L}$ ground truth labels $L_{t}=\left\{l_{0}, l_{1}, l_{2} \ldots, l_{\mathcal{L}-1}\right\}$, in order of decreasing probability $P_{t}=$ $\left[p_{0}, p_{1}, p_{2} \ldots, p_{Q-1}\right].$ The precision at ${k}$ is 
\begin{equation}
P@K=
\frac{1}{k} \sum_{j=0}^{\min (\mathcal{L}, k)-1} {rel}_{L_{i}}\left(P_{t}(j)\right),
\end{equation}
\begin{equation}
\operatorname{rel}_{L}(p)= \begin{cases}1 & \text { if } p \in L \\ 0 & \text { otherwise }\end{cases},
\end{equation}
where $\mathcal{L}$ is the number of ground truth labels or possible answers on each text and $k$ is the number of selected labels on extreme multi-label text classification.

\myparagraph{Normalized Discounted Cummulated Gains (NDCG@K)} The ${NDCG}@{K}$   \cite{DBLP:conf/sigir/LiuCWY17} is 
\begin{equation}
{NDCG} @ K=\frac{1}{{IDCG}\left({L}_{i}, k\right)}\sum_{j=0}^{n-1} \frac{r e l_{L_{i}}\left(P_{t}(j)\right)}{\ln (j+1)},
\end{equation}
where $IDCG$ is ideal discounted cumulative gain and the particular rank position $n$ is
\begin{equation}
n=\min \left(\max \left(\left|P_{i}\right|,\left|L_{i}\right|\right), k\right).
\end{equation}
Among these multi-label evaluation metrics, $ Micro\!-\!\!F1$ considers the number of categories, which makes it suitable for the unbalanced data distribution. $ Macro\!-\!\!F1$ does not take into account the amount of data that treats each class equally. Thus, it is easily affected by the classes with high Recall and Precision. When the number of categories is large or extremely large, either P@K or NDCG@K is used.

\section{Quantitative Results}\label{Section 5}

There are many differences between sentiment analysis, news classification, topic labeling and natural language inference tasks, which can not be simplified modeled as a text classification task. In this section, we tabulate the performance of the main models given in their articles on classic datasets evaluated by classification accuracy, as shown in Table~\ref{tab:Performance}, including MR, SST-2, IMDB, Yelp.P, Yelp.F, Amazon.F, 20NG, AG, DBpedia, and SNLI. 

We give the performance of NB and SVM algorithms from RNTN \cite{DBLP:conf/emnlp/SocherPWCMNP13} due to the less traditional text classification model has been an experiment on datasets in Table~\ref{tab:Performance}. The accuracy of NB and SVM are 81.8\% and 79.4\% on SST-2, respectively.
We can see that, in the SST-2 data set with only two categories, the accuracy of NB is better than that of SVM. It may be because NB has relatively stable classification efficiency on new data sets. The performance is also stable on small data sets. Compared with the deep learning model, the performance of NB is lower. NB has the advantage of lower computational complexity than deep models. However, it requires manual classification features, making it difficult to migrate the model directly to other data sets.

For deep learning models, pre-trained models get better results on most datasets. It means that if you need to implement a text classification task, you can preferentially try pre-trained models, such as BERT, RoBERTa, and XLNET, etc., except MR and 20NG, which have not been experimented on BERT based models. 
Pre-trained models are essential to NLP. It uses a deep model to learn a better feature of the text. It also demonstrates that the accuracy of NLP tasks can be significantly improved by a profound model that can be pre-trained from unlabeled datasets. 
For the MR dataset, the accuracy of RNN-Capsule \cite{DBLP:conf/www/WangSH0Z18} is 83.8\%, obtaining the best result. It suggests that RNN-Capsule builds a capsule in each category for sentiment analysis. It can output words including sentiment trends indicating attributes of capsules with no applying linguistic knowledge. 
For 20NG dataset, BLSTM-2DCNN \cite{DBLP:conf/coling/ZhouQZXBX16} gets 96.5\% score with the best accuracy score. It may demonstrate the effectiveness of applying the 2D max-pooling operation to obtain a fixed-length representation of the text and utilize 2D convolution to sample more meaningful matrix information.

\begin{table*}
    \centering
    \caption{Accuracy of text classification models on primary datasets evaluated by classification accuracy (in terms of publication year). Bold is the most accurate. }\label{tab:Performance}
     \resizebox{\textwidth}{!}{
    \begin{tabular}{l|cccccc|cc|c|c}
    \toprule
        \multirow{2}{*}{\textbf{Model}}&\multicolumn{6}{c|}{\textbf{Sentiment}}&\multicolumn{2}{c|}{\textbf{News}}&\multicolumn{1}{c|}{\textbf{Topic}}&\multicolumn{1}{c}{\textbf{NLI}}\\ 
       
         & MR & SST-2 & IMDB & Yelp.P & Yelp.F & Amz.F & 20NG & AG & DBpedia&SNLI \\ 
        \midrule
         NB \cite{DBLP:journals/jacm/Maron61} &- & 81.80 & - & - & - & - & - & - & -&- \\ \hline
         SVM \cite{DBLP:journals/ml/CortesV95} &- & 79.40 & - & - & - & - & - & - & -&- \\ \hline
        Tree-CRF \cite{DBLP:conf/naacl/NakagawaIK10} &77.30 & - & - & - & - & - & - & - & -&- \\ \hline
        RAE \cite{DBLP:conf/emnlp/SocherPHNM11} & 77.70 & 82.40 & - & - & - & - & - & - & -&- \\ \hline
        MV-RNN \cite{DBLP:conf/emnlp/SocherHMN12} & 79.00 & 82.90 & - & - & - &  -& - &  -& -&- \\ \hline
        RNTN \cite{DBLP:conf/emnlp/SocherPWCMNP13} & 75.90 & 85.40 & - &-  & - &-  &-  &  -& -&- \\ \hline
        DCNN \cite{DBLP:conf/acl/KalchbrennerGB14} &  & 86.80 & 89.40 & - &-  & - & - & - & -&- \\ \hline
        Paragraph-Vec \cite{DBLP:conf/icml/LeM14} &  & 87.80 & 92.58 & - &  -&-  &-  & - & -&- \\ \hline
        TextCNN\cite{DBLP:conf/emnlp/Kim14} & 81.50 & 88.10 & - & - &-  &-  &-  & - &- &- \\ \hline
        TextRCNN \cite{DBLP:conf/aaai/LaiXLZ15} & - &-  &-  &-  & - &-  & 96.49 & - & -&- \\ \hline
        DAN \cite{DBLP:conf/acl/IyyerMBD15} &-  & 86.30 & 89.40 & - & - & - & - & - & - &-\\ \hline
        Tree-LSTM \cite{DBLP:conf/acl/TaiSM15} &  & 88.00 & - &-  & - &-  &-  &-  &- &- \\ \hline
        CharCNN \cite{DBLP:conf/nips/ZhangZL15} &-  & - & - & 95.12 & 62.05 & - & - & 90.49 & 98.45 &-\\ \hline
        HAN \cite{DBLP:conf/naacl/YangYDHSH16} &-  &  -& 49.40 & - & - & 63.60 & - &  -&- &- \\ \hline
        SeqTextRCNN \cite{DBLP:conf/naacl/LeeD16} & - &-  &-  &  -& - & - & - & - & -&- \\ \hline
        oh-2LSTMp \cite{DBLP:conf/icml/JohnsonZ16} & - & - & 94.10 & 97.10 & 67.61 & - & 86.68 & 93.43 & 99.16 &-\\ \hline
        LSTMN \cite{DBLP:conf/emnlp/0001DL16} & - & 87.30 & - &  -& - & - &  -&-  &  -&-\\ \hline
        Multi-Task \cite{DBLP:conf/ijcai/LiuQH16} & - & 87.90 & 91.30 & - &-  & - &-  &-  & -&- \\ \hline
        BLSTM-2DCNN \cite{DBLP:conf/coling/ZhouQZXBX16} & 82.30 & 89.50 &-  &-  &-  & - & \textbf{96.50} &-  & -&- \\ \hline
        TopicRNN \cite{DBLP:conf/iclr/Dieng0GP17} &  -& - & 93.72 & - &  -& - & - &-  & -&- \\ \hline
        DPCNN \cite{DBLP:conf/acl/JohnsonZ17} & - &-  & - & 97.36 & 69.42 & 65.19 & - & 93.13 & 99.12&- \\ \hline
        KPCNN \cite{DBLP:conf/ijcai/WangWZY17} & 83.25 & - &  -&  -&-  &-  &-  & 88.36 & -&- \\ \hline
        RNN-Capsule \cite{DBLP:conf/www/WangSH0Z18} & \textbf{83.80} &  &-  & - &  -& - &-  &  -&- &- \\ \hline
        ULMFiT \cite{DBLP:conf/acl/RuderH18} &  -& - & 95.40 & 97.84 & 71.02 & - &-  & 94.99 & 99.20&- \\ \hline
        LEAM\cite{DBLP:conf/acl/HenaoLCSWWZZ18} & 76.95 &  -&-  & 95.31 & 64.09 & - & 81.91 & 92.45 & 99.02 &-\\ \hline
        TextCapsule \cite{DBLP:conf/emnlp/YangZYLZZ18} & 82.30 & 86.80 &-  & - &-  &-  &-  & 92.60 & -&- \\ \hline
        TextGCN \cite{DBLP:conf/aaai/YaoM019} & 76.74 & - &  -& - &-  & - & 86.34 & 67.61 & -&- \\ \hline
        BERT-base \cite{DBLP:conf/naacl/DevlinCLT19} &  -& 93.50 & 95.63 & 98.08 & 70.58 & 61.60 & - &-  & -&91.00 \\ \hline
        BERT-large \cite{DBLP:conf/naacl/DevlinCLT19} & - & 94.90 & 95.79 & 98.19 & 71.38 & 62.20 & - & - &  -&91.70\\ \hline
        MT-DNN\cite{DBLP:conf/acl/LiuHCG19} & - & 95.60 & 83.20 & - & - & - & - &-  & - &91.50\\ \hline
        XLNet-Large \cite{DBLP:conf/nips/YangDYCSL19} &-  & 96.80 & \textbf{96.21} & \textbf{98.45} & \textbf{72.20} & \textbf{67.74} &-  &-  & -&- \\ \hline
        XLNet \cite{DBLP:conf/nips/YangDYCSL19} &-  & \textbf{97.00} & - &-  &-  & - & - & \textbf{95.51} & \textbf{99.38}&- \\ \hline
        RoBERTa \cite{DBLP:journals/corr/abs-1907-11692} &-  & 96.40 & - &  -&  -& - & - &-  &- &\textbf{92.60} \\ 
        \bottomrule
    \end{tabular}
    }
\end{table*}

\section{Future Research Challenges}\label{Section 6}

Text classification -- as efficient information retrieval and mining technology -- plays a vital role in managing text data. 
It uses NLP, data mining, machine learning, and other techniques to automatically classify and discover different text types. 
Text classification takes multiple types of text as input, and the text is represented as a vector by the pre-training model. Then the vector is fed into the DNN for training until the termination condition is reached, and finally, the performance of the training model is verified by the downstream task. 
Existing models have already shown their usefulness in text classification, but there are still many possible improvements to explore.

Although some new text classification models repeatedly brush up the accuracy index of most classification tasks, it cannot indicate whether the model "understands" the text from the semantic level like human beings. 
Moreover, with the emergence of the noise sample, the small sample noise may cause the decision confidence to change substantially or even lead to decision reversal. 
Therefore, the semantic representation ability and robustness of the model need to be proved in practice. 
Besides, the pre-trained semantic representation model represented by word vectors can often improve the performance of downstream NLP tasks. 
The existing research on the transfer strategy of context-free word vectors is still relatively preliminary. 
Thus, we conclude from data, models, and performance perspective, text classification mainly faces the following challenges.

\subsection{Challenges from Data Perspective}
For a text classification task, data is essential to model performance, whether it is traditional or deep learning method. 
The text data mainly studied includes multi-chapter, short text, cross-language, multi-label, less sample text. 
For the characteristics of these data, the existing technical challenges are as follows:

\myparagraph{Zero-shot/Few-shot learning} 
Zero-shot or few-shot learning for text classification aim to classify text having no or few same labeled class data.
However, the current models are too dependent on numerous labeled data. 
The performance of these models is significantly affected by zero-shot or few-shot learning.
Thus, some works focus on tackling these problems.
The main idea is to infer the features through learning kinds of semantic knowledge, such as learning relationship among classes \cite{DBLP:journals/corr/abs-1712-05972} and incorporating class descriptions \cite{DBLP:conf/naacl/ZhangLG19}.
Furthermore, latent features generation  \cite{DBLP:conf/ijcai/SongZSXX20} meta-Learning \cite{DBLP:conf/emnlp/GengLLZJS19, DBLP:conf/aaai/DengZSCC20, DBLP:conf/iclr/BaoWCB20}
and dynamic memory mechanism \cite{DBLP:conf/acl/GengLLSZ20} are also efficient methods.
Nevertheless, with the limitation of little unseen class data and different data distribution between seen class and unseen class, there is still a long way to go to reach the learning ability comparable to that of humans.

\myparagraph{The external knowledge} 
As we all know, the more beneficial information is input into a DNN, its better performance. For example, a question answering system incorporating a common-sense knowledge base can answer questions about the real world and help solve problems with incomplete information.
Therefore, adding external knowledge (knowledge base or knowledge graph) \cite{DBLP:conf/acl/RojasBOC20, DBLP:journals/mlc/ShanavasWLH21} is an efficient way to promote the model's performance. 
The existing knowledge includes conceptual information \cite{DBLP:conf/ijcai/WangWZY17, DBLP:conf/aaai/ChenHLXJ19, DBLP:journals/corr/abs-1904-09223}, commonsense knowledge \cite{DBLP:conf/emnlp/DingLLLD19}, knowledge base information \cite{DBLP:conf/acl/HaoZLHLWZ17, DBLP:conf/ekaw/TurkerZKS18}, general knowledge graph \cite{DBLP:conf/naacl/ZhangLG19} and so on, which enhances the semantic representation of texts.
Nevertheless, with the limitation of input scale, how and what to add for different tasks is still a challenge.

\myparagraph{Special domain with many terminologies} 
Most of the existing models are supervised models, which over-rely on numerous labeled data. When the sample size is too small, or zero samples occur, the performance of the model will be significantly affected. New data set annotation takes a lot of time. Therefore, unsupervised learning and semi-supervised learning have great potential for text classification. Furthermore, texts in a particular field \cite{DBLP:conf/ijcai/LiangCYLQZ20, DBLP:journals/ijmei/BMHK21}, such as financial and medical texts, contain many specific words or domain experts intelligible slang, abbreviations, etc., which make the existing pre-trained word vectors challenging to work on.

\myparagraph{The multi-label text classification task} 
Multi-label text classification requires full consideration of the semantic relationship among labels, and the embedding and encoding of the model is a process of lossy compression \cite{DBLP:journals/apin/WangLLZZF20, DBLP:journals/kbs/WangHLY21}. 
Therefore, how to reduce the loss of hierarchical semantics and retain rich and complex document semantic information during training is still a problem to be solved.

\subsection{Challenges from Model Perspective}
Most existing structures of traditional and deep learning models are tried for text classification, including integration methods.
BERT learns a language representation that can be used to fine-tune for many NLP tasks. 
The primary method is to increase data, improve computation power, and design training procedures for getting better results \cite{DBLP:conf/coling/DuHM20, DBLP:journals/tcyb/DuVC21, DBLP:journals/corr/abs-2102-00426}.
How the tradeoff between data and compute resources and prediction performance is worth studying. 

\myparagraph{Text representation} 
The text representation method based on the vector space model is simple and effective in the text preprocessing stage. However, it will lose the semantic information of the text, so the application performance based on this method is limited. The proposed semantically based text representation method is too time-consuming. Therefore, the efficient semantically based text representation method still needs further research.
In the text representation of text classification based on deep learning, word embedding is the main concept, while the representation unit is described differently in different languages. Then, a word is represented in the form of a vector by learning mapping rules through the model. Therefore, how to design adaptive data representation methods is more conducive to the combination of deep learning and specific classification tasks.

\myparagraph{Model integration} 
Most structures of traditional and deep learning models are tried for text classification, including integration methods.  RNN requires recursive step by step to get global information. CNN can obtain local information, and the sensing field can be increased through the multi-layer stack to capture more comprehensive contextual information. Attention mechanisms learn global dependency among words in a sentence. The transformer model is dependent on attention mechanisms to establish the depth of the global dependency relationship between the input and output. Therefore, designing an integrated model is worth trying to take advantage of these models.

\myparagraph{Model efficiency} 
Although text classification models based on deep learning are highly effective, such as CNNs, RNNs, and GNNs. However, there are many technical limitations, such as the depth of the network layer, regularization problem, network learning rate, etc. Therefore, there is still more broad space for development to optimize the algorithm and improve the speed of model training.

\subsection{Challenges from Performance Perspective}
The traditional model and the deep model can achieve good performance in most text classification tasks, but the anti-interference ability of their results needs to be improved \cite{DBLP:conf/emnlp/ZhouJCW19, DBLP:conf/aaai/JinJZS20, DBLP:journals/corr/abs-2103-04264}. 
How to realize the interpretation of the deep model is also a technical challenge.

\myparagraph{The semantic robustness of the model} 
In recent years, researchers have designed many models to enhance the accuracy of text classification models. 
However, when there are some adversarial samples in the datasets, the model's performance decreases significantly. Adversarial training is a crucial method to improve the robustness of the pre-training model. For example, a popular approach is converting attack into defense and using the adversarial sample training model. 
Consequently, how to improve the robustness of models is a current research hotspot and challenge.

\myparagraph{The interpretability of the model} 
DNNs have unique advantages in feature extraction and semantic mining and have achieved excellent text classification tasks. Only a better understanding of the theories behind these models can accurately design better models for various applications. 
However, deep learning is a black-box model, the training process is challenging to reproduce, and the implicit semantics and output interpretability are poor. 
It makes the improvement and optimization of the model, losing clear guidelines. 
Why does one model outperform another on one data set but underperform on others? What does the deep learning model learn? 
Furthermore, we cannot accurately explain why the model improves performance.

\section{Conclusion}\label{Section 7}
This paper principally introduces the existing models for text classification tasks from traditional models to deep learning. 
Firstly, we introduce some primary traditional models and deep learning models with a summary table. 
The traditional model improves text classification performance mainly by improving the feature extraction scheme and classifier design. 
In contrast, the deep learning model enhances performance by improving the presentation learning method, model structure, and additional data and knowledge.
Then, we introduce the datasets with a summary table and evaluation metrics for single-label and multi-label tasks. 
Furthermore, we give the quantitative results of the leading models in a summary table under different applications for classic text classification datasets. 
Finally, we summarize the possible future research challenges of text classification.

\begin{acks}
The authors of this paper were supported by the National Key R\&D Program of China through grant 2021YFB1714800, NSFC through grants (No.U20B2053 and 61872022), State Key Laboratory of Software Development Environment (SKLSDE-2020ZX-12).
Philip S. Yu was supported by NSF under grants III-1763325, III-1909323, III-2106758, and SaTC-1930941.
Lifang He was supported by NSF ONR N00014-18-1-2009 and Lehigh's accelerator grant S00010293.
This work was also sponsored by CAAI-Huawei MindSpore Open Fund. 
Thanks for computing infrastructure provided by Huawei MindSpore platform.
\end{acks}

\footnotesize
\bibliographystyle{ieeetr}
\bibliography{TextClassification}

\appendix

\end{document}